\documentclass[10pt,twocolumn,letterpaper]{article}

\usepackage[pagenumbers]{cvpr} 
\usepackage{colortbl}
\usepackage{multirow}
\usepackage{booktabs}
\usepackage{pifont}%
\usepackage[accsupp]{axessibility}
\usepackage{pgfplots}
\usepackage[accsupp]{axessibility}
\newcommand\bench{{BlackSwanSuite}}
 \usepackage{multirow}
\definecolor{cvprblue}{rgb}{0.21,0.49,0.74}
\usepackage[pagebackref,breaklinks,colorlinks,allcolors=cvprblue]{hyperref}
\usepackage{multirow}
\usepackage{caption}
\usepackage{algorithm}
\usepackage{algpseudocode}

\newcommand{\code}[2]{%
  \begingroup\setlength{\fboxsep}{1pt}%
  \colorbox{#1}{#2}%
  \endgroup
}
\newcommand{\taskOne}{\code{pink!20}{Forecaster}}
\newcommand{\taskTwo}{\code{green!10}{Detective}}
\newcommand{\taskThree}{\code{cyan!10}{Reporter}}

\usepackage{verbatim}
\usepackage{multicol}
\newcommand{\BlackSwanImage}{\includegraphics[scale=0.009]{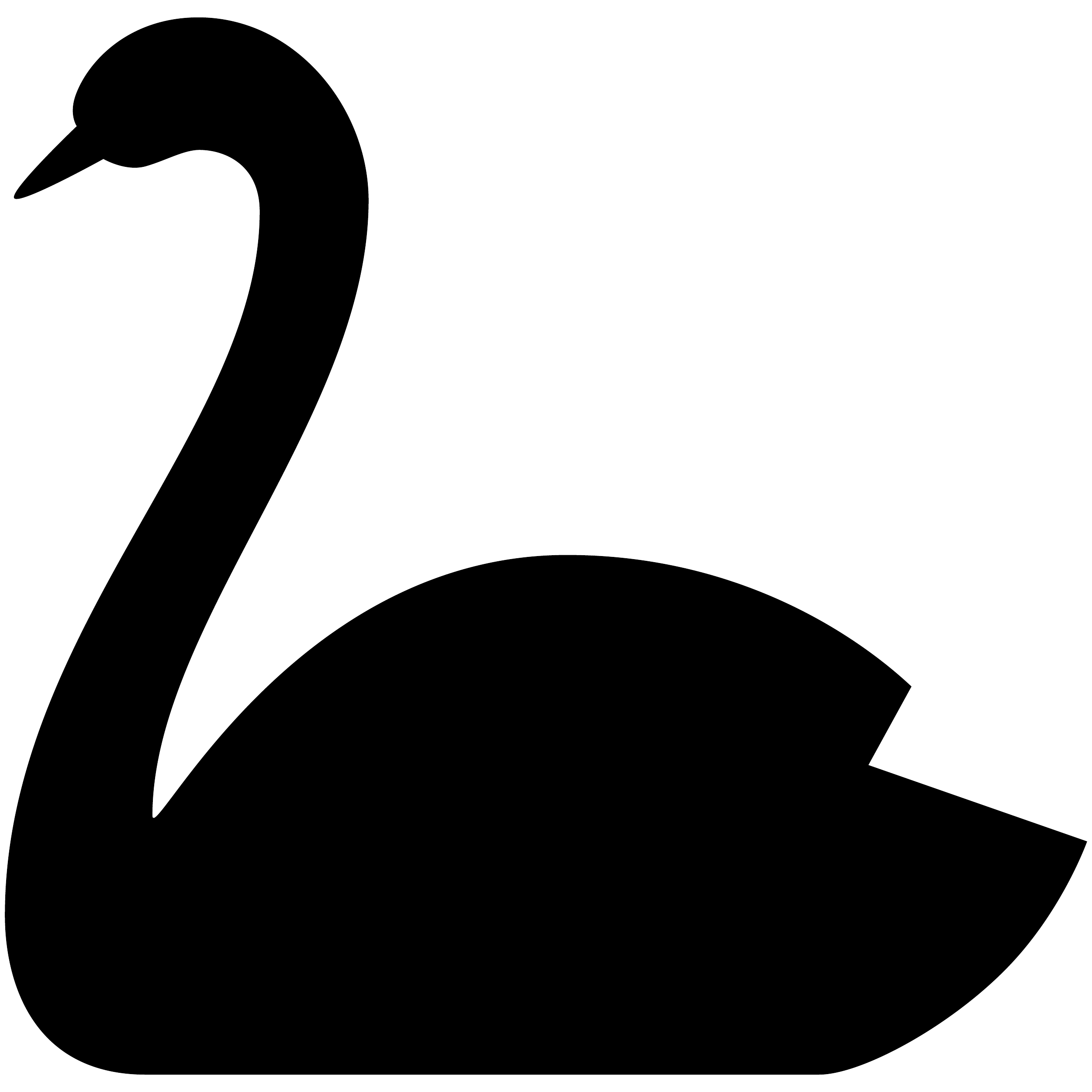}\hspace{0.1cm}}%

\newcommand{\abl}[1]{{\color{red}[{#1}]$_{-ABL}$}}

\definecolor{green}{rgb}{0,0.6,0}
\definecolor{red}{rgb}{0.8, 0.25, 0.33}

\def\paperID{14335} 
\def\confName{CVPR}
\def\confYear{2025}

\title{\BlackSwanImage {Black Swan}: Abductive and Defeasible Video Reasoning \\ in Unpredictable Events}

\author{Aditya Chinchure\thanks{Denotes Equal Contribution} $^{1,2}$
~~~~~
Sahithya Ravi$^{*1,2}$
~~~~~
Raymond Ng$^1$
~~~~~
Vered Shwartz$^{1,2}$ \\
~~~~~
Boyang Li$^3$  
~~~~~
Leonid Sigal$^{1,2}$\\
~~
$^1$University of British Columbia ~~~ $^2$Vector Institute for AI ~~~ $^3$Nanyang Technological University \\
{\tt\small \{aditya10, sahiravi, rng, vshwartz, lsigal\}@cs.ubc.ca} ~~~~~
~~~~~
{\tt\small boyang.li@ntu.edu.sg}
}
\begin{document}
\maketitle
\begin{abstract}

The commonsense reasoning capabilities of vision-language models (VLMs), especially in abductive reasoning and defeasible reasoning, remain poorly understood.
Most benchmarks focus on typical visual scenarios \cite{Agrawal2015VQAVQ,zellers2019vcr, park2020visualcomet}, making it difficult to discern whether model performance stems from keen perception and reasoning skills, or reliance on pure statistical recall. We argue that by focusing on atypical events in videos, clearer insights can be gained on the core capabilities of VLMs. Explaining and understanding such out-of-distribution events requires models to extend beyond basic pattern recognition and regurgitation of their prior knowledge. To this end, we introduce \bench{}, a benchmark for evaluating VLMs' ability to reason about unexpected events through abductive and defeasible tasks. Our tasks artificially limit the amount of visual information provided to models while questioning them about hidden unexpected events, or provide new visual information that could change an existing hypothesis about the event. We curate a comprehensive benchmark suite comprising over 3,800 MCQ, 4,900 generative and 6,700 yes/no questions, spanning 1,655 videos. After extensively evaluating various state-of-the-art VLMs, including GPT-4o and Gemini 1.5 Pro, as well as open-source VLMs such as LLaVA-Video, we find significant performance gaps of up to 32\% from humans on these tasks. Our findings reveal key limitations in current VLMs, emphasizing the need for enhanced model architectures and training strategies. Our data and leaderboard is available at \url{https://blackswan.cs.ubc.ca}.

\end{abstract}

\section{Introduction}
\label{sec:intro}
\begin{figure*}
  \centering
  \includegraphics[width=1.0\linewidth]{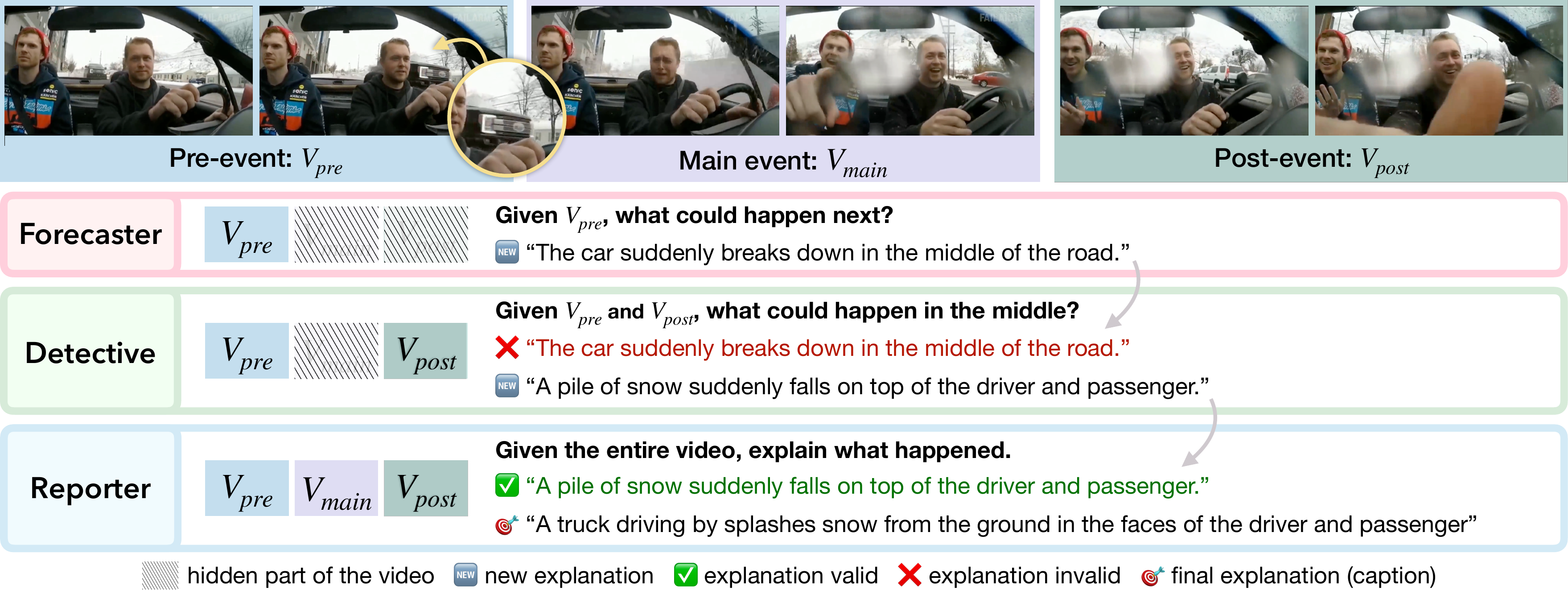}
   \caption{\textbf{\bench{}}. Our benchmark involves three tasks: i) \taskOne{} evaluates a model's ability to hypothesize future events. ii) \taskTwo{} involves abductive reasoning by explaining the hidden event, and defeasible reasoning by validating existing hypotheses.  iii) \taskThree{} again tests defeasability and the model's ability to describe the unexpected event.}
   \label{fig:tasks}
   \vspace{-0.2cm}
\end{figure*}

Vision-language models (VLMs) are becoming increasingly capable of reasoning about the world thanks to their exposure to vast amounts of visual data, and the emergent capabilities of their underlying large language models (LLMs). Recent multi-frame video-language models such as VILA \cite{lin2023vila}, Video-LLaMA \cite{cheng2024videollama2}, and LLaVA-Video \cite{zhang2024llavavideo} show promising results in tasks such as video captioning and question answering. However, it remains unclear to what extent they can reason about 
unexpected events. 

Unexpected events pose a unique challenge to AI models, as they deviate from the patterns in the training set, leaving the models less equipped to handle them \cite{role-of-memorization-2017,pmlr-v97-rahaman19a,Dimitris2019}. However, traditional benchmarks for video reasoning primarily focus on predictable scenarios ({\em e.g.} \cite{Xu2017VideoQA}), overlooking the critical challenge of assessing how models handle rare, unforeseen events. Furthermore, these benchmarks often do not require models to revise their reasoning in response to conflicting or new evidence.

The human ability to understand, rationalize, and respond to unexpected events is underpinned by \emph{abductive reasoning}, or drawing the most likely explanation from limited observations, and \emph{defeasible reasoning}, where initial conclusions are revised in light of new or conflicting evidence. For example, a human may observe two damaged cars in an intersection, and hypothesize the cause as one driver running the red light (\textit{abductive}). Later, on observing that the traffic lights malfunctioning, they instead believe the cause was faulty traffic lights rather than the driver (\textit{defeasability}). If AI models are to function as autonomous decision makers ({\em e.g.}, in self-driving vehicles \cite{OKelly2018,Wang2021AdvSim}), reasoning about unexpected events abductively and defeasibly would be critical to their safety and real-world utility.

Motivated by the need to evaluate the gap in abductive and defeasible reasoning abilities between humans and models, and the limitations of existing benchmarks, we propose the novel \textbf{\bench{}} benchmark. Our benchmark contains a range of tasks that require nuanced perception, comprehension, and reasoning abilities. We focus on leveraging the visual information provided in videos of expectation-violating events, including but not limited to surprises, accidents, pranks and other uncommon situations. Such videos are ideal for evaluating abductive and defeasible reasoning capabilities of models, as these events occur scarcely, if ever, in the training data, and regurgitation of seen data arguably would be  insufficient. Our benchmark consists of three overarching tasks (Fig. \ref{fig:tasks}), named \taskOne, \taskTwo, and \taskThree, involving 15,469 questions, with both generative and discriminative variants. 

We evaluate leading closed-source and open-source video VLMs and multi-frame VLMs on our tasks. Results indicate that the best models lag behind humans by up to 25\% on Multiple Choice Questions (MCQ) and 32\% on Yes/No (Y/N) questions.
By highlighting the limitations of current VLMs on reasoning tasks, \bench{} can drive the development of future models with stronger reasoning abilities.






\section{Background}
\label{sec:related_work}

\bench{} focuses on evaluating models on their ability to perform abductive and defeasible reasoning in videos containing unpredictable scenarios. We first provide background on abductive reasoning (\S\ref{sec:bg:abd}) and defeasible reasoning (\S\ref{sec:bg:def}) along with related work in this area. We then describe related video language benchmarks (\S\ref{sec:bg:videobench}).

\subsection{Abductive Reasoning}
\label{sec:bg:abd}
Abductive reasoning focuses on finding the most plausible explanation $H$ for a set of observations 
$\mathcal{O} = \{O_1$,\ldots, $O_n \}$~\cite{peirce1974collected}. 
Consider the following example:
\begin{itemize}
    \item \textbf{Observation 1 ($O_1$)}: ``The door was left open.''
    \item \textbf{Observation 2 ($O_2$)}: ``A broken vase lay on the floor.''
\end{itemize}

\noindent The following hypothesis $H$ is a plausible explanation for what might have happened between $O_1$ and $O_2$:

\begin{itemize}
    \item \textbf{Hypothesis ($H$)}: ``A cat entered through the open door and knocked over the vase.''
\end{itemize}

\noindent Though other explanations exist ({\em e.g.}, ``a black swan flew in and knocked over the vase''), abductive reasoning favors the most likely hypothesis based on typical scenarios.


Computational abductive reasoning has drawn interest in NLP, with Bhagavatula et al. \cite{Bhagavatula2019AbductiveCR} initiating a task to generate plausible explanations for narratives. Du et al. \cite{du2022graph} and Paul and Frank \cite{paul2021generating} extended this work, using graph-based methods and models predicting event sequences. Qin et al. \cite{qin2020back} explored abductive and counterfactual reasoning, while Liu et al. \cite{liu2023revisiting} emphasized unexpected scenarios. Earlier, Ovchinnikova et al. \cite{Ovchinnikova2011AbductiveRW} addressed abductive reasoning in discourse parsing.

In the vision-language domain, Sherlock \citep{Hessel2022TheAO} provides a visual abductive dataset where models rank 10 inference candidates, aligning with human preference on inferred explanations. \citet{Liang2022VisualAR} introduced causality in prediction tasks by generating explanation events based on premises. VideoABC \citep{Zhao2022VideoABCAR} framed an abductive reasoning task tailored to instructional videos, where the model must infer likely sequences of keyframes.

Our work differs from the prior approaches \citep{Liang2022VisualAR, Zhao2022VideoABCAR} by focusing explicitly on unexpected events and measuring defeasibility. This task requires reasoning grounded in visual content rather than relying solely on language cues, marking a distinct shift in the abductive reasoning landscape.

\subsection{Defeasible Reasoning}
\label{sec:bg:def}
Defeasible reasoning is a form of reasoning where conclusions are drawn tentatively, allowing for revision if conflicting evidence arises \cite{reiter1980logic}. Formally, let $P$ represent initial premises and let $C$ be a defeasible conclusion drawn from $P$. If new information $P'$ is introduced, it may lead to $\neg C$ (i.e., the invalidation of $C$), thereby ``defeating'' the initial conclusion.
For example, 
given the initial premises $P$, we might infer $C$. 



\begin{itemize}
    \item \textbf{Initial Premise ($P$)}: ``The door was open and a valuable
vase was missing from its stand.'' 
    \item \textbf{Conclusion ($C$)}: ``A valuable vase was stolen.''
\end{itemize}
However, if we add new context:

\begin{itemize}
    \item \textbf{New Evidence ($P'$)}: ``The vase, broken into pieces, lay on the floor.''
\end{itemize}
The conclusion $C$ is now defeated, and we can no longer infer that the vase was stolen.

Early defeasibility work in language tasks includes Rudinger et al. \cite{rudinger2020thinking} on premise updates and effect classification, and Madaan et al. \cite{Madaan2021ThinkAI} with inference graphs for nuanced reasoning. Zhou et al. \cite{Zhou2023COBRAFC} used Contextual Bias Frames for bias and moral reasoning tasks \cite{Rao2023WhatMI}. Recently, Cui et al. \cite{Cui2024CAUSALED} introduced a defeasibility dataset for causal reasoning. Some work has also considered defeasibility in the visual realm; Han et al. \cite{han2023reading} investigated defeasibility of social norms using images as visual context. However, no prior work has addressed defeasibility in the context of unexpected events, nor has any explored defeasible reasoning across video or multi-frame formats. Our benchmark, \bench{}, introduces the first task of this nature.

\subsection{Foundational Vision Language Reasoning}
\label{sec:bg:videobench}
Traditional benchmarks in vision and language reasoning focus on reasoning about commonly occurring video events. 
For instance, early video reasoning benchmarks such as TGIF-QA \cite{Jang2017TGIFQATS} and ActivityNet-QA \cite{Yu2019ActivityNetQAAD} challenge models to answer questions involving spatio-temporal reasoning. Moving beyond this, CLEVRER \cite{Yi2019CLEVRERCE} and NExT-QA \cite{Xiao2021NExTQANP} introduce causal and counterfactual reasoning, while Causal-VidQA \cite{li2022from} emphasizes causal inference within video content.
Human-centered reasoning benchmarks, such as VCR \cite{zellers2019vcr} and MovieQA \cite{tapaswi2016movieqaunderstandingstoriesmovies}, further test multi-modal understanding by requiring models to interpret social cues in videos. These benchmarks assess sophisticated reasoning within human-centric environments, though they are typically limited to normative situations without unexpected elements. Some benchmarks specifically target surprising scenarios. For example, FunQA \cite{Xie2023FunQATS} asks questions about funny videos. However, these questions do not specifically involve abduction or defeasibility, and target video captioning and conversation instead.



\section{Tasks}
\label{sec:tasks}


Surprising or unexpected events often follow a structured narrative, beginning with a normal scenario, followed by an unexpected event that deviates from the norm, and concluding with an unlikely outcome. We leverage this narrative structure of our videos. As shown in Figure~\ref{fig:tasks}, each video is divided into three parts: the \emph{pre-event} (\( V_{pre} \)), showing the premise, or the events leading up to the unexpected event; the \emph{main event} (\( V_{main} \)), where the unexpected event occurs; and the \emph{post-event} (\( V_{post} \)), which reveals the outcome of the main event and concludes the video. \bench{} has three tasks based on the amount of available video information to a model, each testing different reasoning abilities:

\subsection{\taskOne{}: Predict the Future}
In this task, models are only shown the pre-event, \( V_{pre} \), and asked to predict the next event. This tests the model's ability to evaluate the scenario in the video, and explain future trajectories.
This task only contains one variant:
\begin{itemize}
    \item \textbf{\taskOne--Gen}: Generate a free-text answer to ``What happens next?"
\end{itemize}

\subsection{\taskTwo{}: Explain the Outcome}
This task presents models with \( V_{pre} \) and \( V_{post} \) and asks them to reason about what could have happened in-between, i.e., in the \emph{main event} \( V_{main} \), requiring abductive reasoning. Furthermore, this task tests the defeasible reasoning ability of the model by asking it to validate or invalidate a hypothesis of what could be happening in \( V_{main} \). This task has three variants:

\begin{itemize}
    \item \textbf{\taskTwo--Gen}: Generate a free-text answer to ``What happened in the middle?"
    \item \textbf{\taskTwo--MCQ}: Choose one of three options for ``What happened in the middle?"
    \item \textbf{\taskTwo--Y/N}: Validate a previous hypothesis about \( V_{main} \).
\end{itemize}

\subsection{\taskThree{}: Describe All Events}
In this task, models see \( V_{pre} \), \( V_{main} \), and \( V_{post} \) (the entire video) and are asked to describe the entire sequence of events. In addition, it tests defeasability by asking models to validate or invalidate a previous hypothesis using the context provided by the entire video. This task has three variants: 

\begin{itemize}
    \item \textbf{\taskThree--Gen}: Generate a free-text explanation of the entire video.
    \item \textbf{\taskThree--MCQ}: Choose the best description of the video's events.
    \item \textbf{\taskThree--Y/N}: Confirm if a hypothesis about \( V_{main} \) holds with full context.
\end{itemize}

\noindent\textbf{Reasoning Types.} \taskTwo{} assesses \emph{abductive reasoning}, requiring models to infer the most plausible cause of the post-event ($V_{post}$) given pre-event context ($V_{pre}$). Both \taskTwo{} and \taskThree{} test \emph{defeasible reasoning}, through in MCQ and Y/N formats, where models evaluate descriptions given new video context. All tasks also test commonsense reasoning capabilities (Appendix \ref{app:reasoningtypes}).

\section{The \bench\ Dataset}
\label{sec:data}
The data collection process for \bench\ dataset summarized in Fig. \ref{fig:datacollection} in Appendix \ref{app:DataCollection}. Below, we describe the source and types of videos in our dataset (\S\ref{sec:data:videos}) and the annotation process (\S\ref{sec:data:annotation}), the creation of the task variants (\S\ref{sec:data:taskvariants}), and the dataset statistics (\S\ref{sec:data:stats}). 


\begin{figure*}
  \centering
  \includegraphics[width=0.9\linewidth]{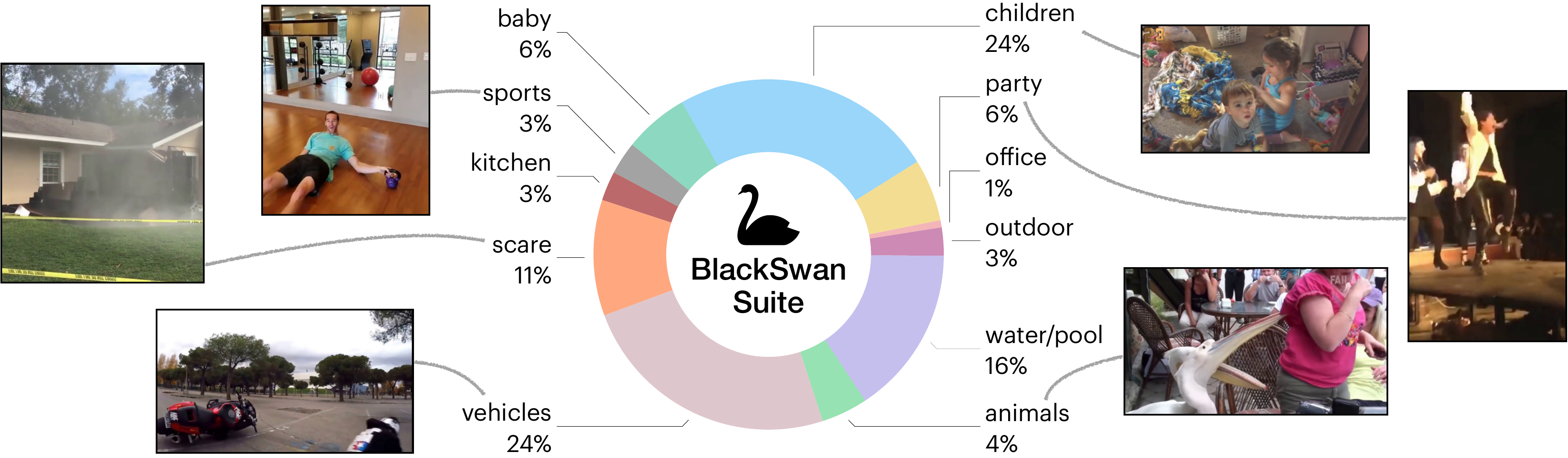}
   \caption{{\bf \bench{}} contains 1655 videos from variety of topics, as depicted above.}
   \label{fig:topics}
\end{figure*}

\subsection{Videos}
\label{sec:data:videos}

\noindent\textbf{Source.} Videos in \bench\ are short clips that contain one surprising event. We obtain the videos from the \textit{test set} of the Oops! dataset \cite{epstein2020oops}, which consists of YouTube fail videos along with localization annotations for the main event occurring in the video. We filter out videos for which there was poor inter-annotator agreement on the localization, or the video contained multiple scenes. 
\\
\noindent\textbf{Splitting to parts.} We divide each video into the three parts defined in Sec.~\ref{sec:tasks}: \emph{pre-event} ($V_{pre}$), \emph{main event} ($V_{main}$), and \emph{post-event} ($V_{post}$). We use the provided localization annotation to identify the main event, and use a combination of an automatic scene splitter and heuristics, as described in Appendix~\ref{app:DataCollectionSplitAlgo} to obtain the three parts of the video. Following the filtering criterion and the splitting process, we have 1655 videos, each with three parts that are at least one second long.

\subsection{Annotation Process}
\label{sec:data:annotation}

We collected annotations for the three tasks defined in Sec.~\ref{sec:tasks}. The annotation task was done in three corresponding steps. In the first step, we showed annotators only the pre-event $V_{pre}$ and asked them to come up with three possible scenarios for what could happen next (\taskOne{}). 

In the second step, we revealed the post-event $V_{post}$, and asked annotators the abductive question ``What could have happened in the middle part of the video?'' (\taskTwo). They were  asked to first validate or invalidate their responses to the first step, and then to write new responses to the ones they invalidated.

In the final step, we revealed the main event $V_{main}$. At this stage, the entire video was visible. Again, we asked annotators to validate or invalidate their answers for the second step. Finally, we asked them to write an explanation of what happened, much like a caption describing the unexpected events in the video (\taskThree).

Our annotation process was conducted through the CloudConnect Platform by Cloud Research. Each video was annotated by a single qualified annotator who was compensated \$0.85 per annotation task, which we estimate sums up to \$10.2 per hour. We further filtered out the worker pool following the validation of 10\% of the collected data by one of the authors, annotating the rest of the data by workers who were adept in the task. 



\noindent\textbf{Data Quality Validation.} We conduct a user study to validate the dataset quality. Details of the user study setup are in Appendix \ref{app:DataCollectionQValid}. We measure correctness of the responses to the questions in the task, depth in reasoning, and a reasonable level of grammatical correctness of 60 randomly sampled videos and all their tasks. Our user study shows that our dataset achieved ratings of 9.6/10 on correctness, 9.3/10 on depth and 92.3\% on grammar. Feedback suggested that the minor mistakes could be attributed to variance in how different people may interpret the same scenario.

\subsection{Task Variants}
\label{sec:data:taskvariants}

Using the annotations, we build three variants of tasks, as described below.
\noindent\textbf{Generative (Gen):}  Every question in \taskOne--Gen comes with 3  ground truth hypotheses that were proposed by annotators in step 1. Questions in \taskTwo--Gen also come with 3 ground truth hypotheses which include valid hypotheses from step 1 and new hypotheses collected in step 2. Finally, \taskThree--Gen has a single reference which is the caption collected from annotators in the last step.

\noindent\textbf{MCQ:} Each MCQ for \taskTwo{} has 3 choices. For the correct choice, we used explanations from step 1 (\taskOne{}) that were validated in step 2 as well as answers for step 2 (\taskTwo{}), duplicating the generative question into multiple MCQs where multiple correct answers were available. For the distractors we used hypotheses from step 1 that were invalidated in step 2 after observing the new information in $V_{post}$. For questions in which we had fewer than 3 incorrect answers, we generated a caption of $V_{pre}$ using a VLM (LLaVA-Video \cite{zhang2024llavavideo}) and used an LLM (GPT-4o \cite{openai2024gpt4o}) to edit it to match the style of the other incorrect options. We followed the same process for \taskThree--MCQ as well, where the correct option is the caption written by the annotator in step 3 or the validated hypothesis from step 2, and the wrong choices are either the invalidated hypothesis, or the $V_{pre}$ caption.

\noindent\textbf{Y/N:} The Y/N variants for \taskOne{} (\taskTwo{}) include each hypothesis proposed in step 1 (step 2) along with its ground truth annotation indicating whether it was validated (yes) or invalidated (no) in step 2 (step 3) on the basis of new visual evidence. The Y/N variant differs from the MCQ variant as it directly tests each hypothesis through defeasible reasoning, rather than evaluating one hypothesis against another as with the options in the MCQ.

\subsection{Dataset Statistics}
\label{sec:data:stats}

\begin{figure}
  \centering
  \includegraphics[width=0.9\linewidth]{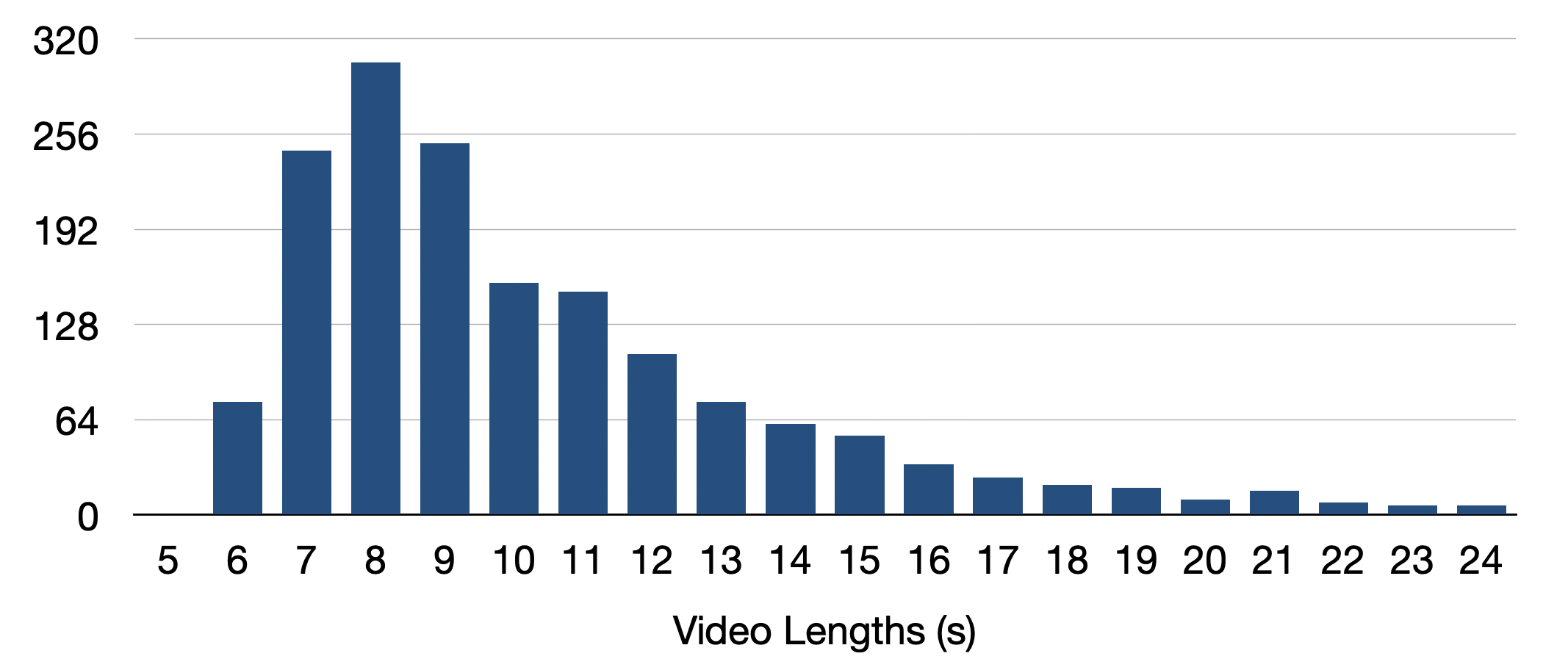}
   \caption{\textbf{Length of Videos.} The median video length is 8.83 seconds. Only a small number of videos are outliers, with 29 of them being longer than 25 seconds.}
   \label{fig:videolen}
\end{figure}

\bench\ contains 1,655 videos from a wide range of topics (Fig~\ref{fig:topics}), ranging from vehicle or road accidents, through children videos, to pranks and scare clips. Figure~\ref{fig:videolen} (Appendix \ref{app:DataCollection}) shows the distribution of video lengths, where the median video length is 8.83 seconds. Only 29 videos have a length greater than 25 seconds.

\begin{table}[ht]
  \centering
  \begin{tabular}{lccc}
    \toprule
    \textbf{Tasks:} & \textbf{Generative} &  \textbf{MCQ} & \textbf{Yes/No} \\
    \midrule
    \taskOne{}  & 1,655 & - & - \\
    \taskTwo{} & 1,655 & 2,415 & 4,917 \\
    \taskThree{} & 1,655 & 1,410 & 1,762 \\
    \bottomrule
  \end{tabular}
  \caption{\textbf{\bench{}} contains 15,469 questions across three tasks and three variants.}
  \label{tab:datasetinfo}
  \vspace{-0.2cm}
\end{table}

In Table \ref{tab:datasetinfo}, we summarize the number of tasks in \bench. Among all the MCQ questions, 1048 have been augmented using the captioning process for a wrong choice.


\section{Experimental Setup}
\label{sec:exp_setup}
We evaluate the performance (\S\ref{sec:exp_setup:eval_metrics}) of various baselines (\S\ref{sec:exp_setup:baselines}) on \bench{}.

\subsection{Evaluation Metrics}
\label{sec:exp_setup:eval_metrics}

We report models' accuracy on the MCQ and Y/N variants. The quality of outputs generated for the generative variants of the tasks is evaluated using a combination of CLIP-based and LLM-based metrics and human evaluation. Given the open-ended nature of \taskOne{} and \taskTwo{}, we generate 3 responses for each question from each model. For \taskThree{}, we only generate a single explanation, since the entire video is revealed to the model. 

\noindent\textbf{CLIP Score.} We embed each model-generated response and each reference explanation in CLIP \cite{radford2021clip}, and compute a pair-wise similarity score. We report the maximum pair-wise similarity for each question, since we want to reward models for coming up with \emph{any} plausible explanation. 

\noindent\textbf{LLM-Match.} Inspired by OpenEQA \cite{OpenEQA2023}, we prompt the LLM to rate the similarity between two sentences on a scale of 1-5, providing it with every pair of reference and system-generated explanation (the full prompt is given in Appendix \ref{app:Metrics}). We compute the average\footnote{We compute average instead of max because we want to measure overall quality for all responses. Please see Appendix \ref{app:Metrics} for details.} similarity score across all pairs. We use Llama 3.1 8B \cite{dubey2024llama3herdmodels} for this process, since it is open source and enables reproducibility.

\noindent\textbf{Human Evaluation.} A detailed description of the human evaluation setup, including the template, is in Appendix \ref{app:HumanEval}. In summary, we ask humans to evaluate a generated response on four parameters: \textbf{C}orrectness (between 1-5, rate how well does it answer the task question), Depth and \textbf{T}houghtfulness (between 1-5, rate the thoughtfulness of the sentence), Level of \textbf{D}etail (between 1-5, rate how well does it describe the scene, the people/objects and actions), and \textbf{V}isual Entailment (0 or 1, is the description possible w.r.t. the video shown). 

\subsection{Baselines}
\label{sec:exp_setup:baselines}

Our evaluation encompasses both open-source and closed-source VLMs. In general, these models may be Video LMs (where the input is a video file, and the frames are sampled by the model) or multi-frame VLMs (where we directly provide uniformly sampled frames). We attempt to test the latest variants of these models. Our baselines include OpenAI's GPT-4o \cite{openai2024gpt4o} and Google's Gemini 1.5 Pro \cite{team2024gemini}, both leading closed-source VLMs with video understanding capabilities. Furthermore, among open-source models, we test LLaVA-Video \cite{zhang2024llavavideo} (latest in LLaVA-Next series), VILA \cite{lin2024vila}, VideoChat2 \cite{li2024mvbenchvideochat2}, and VideoLLaMA 2 \cite{cheng2024videollama2}. These models have shown competitive performance on benchmarks such as MLVU \cite{MLVU}. Specific details about each model, including the prompts used for each task, and the variants of each model used are shown in Appendix \ref{app:Baselines}.

Finally, we also report human performance. For MCQ and Y/N variants of \taskTwo{} and \taskThree{}, we ask a human expert 150 questions for each task variant. For the generative variant, we crowd source human annotations for 20 videos, across all three tasks.



\section{Results}
\label{sec:results}
We show results on all tasks in (\S\ref{sec:analysis:main}) to compare model and human performance, and show qualitative results (\S\ref{sec:analysis:qualitative}). Additional information on data release and leaderboard results are in Appendix \ref{app:leaderboard}.

\subsection{Main Results}
\label{sec:analysis:main}

\setlength{\tabcolsep}{3pt} 
\begin{table}[t]
\centering
\begin{tabular}{lcccc}
\toprule
\multicolumn{1}{c}{\multirow{2}{*}{\textbf{Model}}} & \multicolumn{2}{c}{\textbf{\taskTwo}} & \multicolumn{2}{c}{\textbf{\taskThree}} \\
\multicolumn{1}{c}{} & \textbf{MCQ} & \textbf{Y/N} & \textbf{MCQ} & \textbf{Y/N} \\
\midrule
\rowcolor{gray!20} \multicolumn{5}{c}{\textbf{Closed Source}} \\
GPT-4o & \underline{65.1} & 62.4 & \underline{79.3} & \underline{60.1} \\
Gemini 1.5 Pro & 58.7 & \underline{63.2} & 71.0 & 52.9 \\
\rowcolor{gray!20} \multicolumn{5}{c}{\textbf{Open Source}} \\
VideoChat2 & 29.9 & 63.0 & 51.6 & 44.3 \\
VideoLLaMA 2 & 53.5 & 57.7 & 53.2 & 54.1 \\
VILA-1.5 & 51.8 & 57.6 & 54.3 & 50.6 \\
LLaVA-Video & 55.9 & 59.3 & 69.6 & 55.1 \\
\rowcolor{gray!20} \multicolumn{5}{c}{\textbf{Human}} \\
Human & \textbf{90.0} & \textbf{85.3} & \textbf{95.3} & \textbf{92.0} \\
\bottomrule
\end{tabular}
\caption{\textbf{Results on MCQ and Y/N variants of \taskOne and \taskTwo.} The best result for each task is shown in bold, and the best model result is underlined.}
\label{tab:disc}
\end{table}

\subsubsection{MCQ and Y/N}
In Table~\ref{tab:disc}, we present the results for both the MCQ and Y/N variants of \taskTwo{} and \taskThree{}. In \taskTwo's MCQ and Y/N tasks, we observe a notable performance gap, with humans outperforming the best model, GPT-4o, by 24.9\% on the abductive reasoning MCQ task and by 26\% on the defeasible reasoning Y/N task. These tasks challenge models not only due to limited exposure to such video-based reasoning but also due to the nuanced perception and comprehension required to succeed in abductive and defeasible reasoning (see \S\ref{sec:analysis:perception_comp_reason}). Among closed-source models, GPT-4o achieves the highest performance on both tasks, while LLaVA-Video leads the open-source models. VideoChat2 demonstrates the weakest performance the MCQ task, partly because 35.4\% responses could not be parsed as one of the MCQ options.

As in \taskTwo, models struggle with both the MCQ and Y/N questions in \taskThree{}, where performance lags significantly behind human results. For example, GPT-4o trails humans by 21\% on the MCQ task and 32\%, on the Y/N defeasible task, a gap largely due to our tasks requiring a deeper visual understanding of the scene. Many MCQ questions hinge on specific actions and subtle behaviors of individuals, which can confound models. While they may grasp the general scene, models often struggle to discern the finer details needed to accurately evaluate MCQ options or validate the hypothesis in the Y/N variant. Results vary across the two variants, however, as with the MCQ questions, models may prefer one answer over the others due to stylistic variations or word choice between different options, whereas with the Y/N variant, they judge each answer individually.

\setlength{\tabcolsep}{2pt}
\begin{table}[t]
  \centering
  \begin{tabular}{lcccccc}
    \toprule
    \multicolumn{1}{c}{\multirow{2}{*}{\textbf{Model}}} & \multicolumn{2}{c}{\textbf{Automatic}} & \multicolumn{4}{c}{\textbf{Human Rating}} \\
    \multicolumn{1}{c}{} & \multicolumn{1}{c}{\textbf{CLIP}} & \multicolumn{1}{c}{\textbf{LLM-M}} & \textbf{C} & \textbf{T} & \textbf{D} & \textbf{V} \\
    \midrule
    \rowcolor{gray!20} \multicolumn{7}{c}{\textbf{Closed Source}} \\
    GPT-4o & 0.77 & 1.64 & \textbf{\underline{3.72}} & \textbf{\underline{3.62}} & \textbf{\underline{3.81}} & \underline{0.91} \\
    Gemini 1.5 Pro & \textbf{\underline{0.78}} & 1.61 & 3.30 & 2.90 & 2.95 & 0.85 \\
    \rowcolor{gray!20} \multicolumn{7}{c}{\textbf{Open Source}} \\
    VideoChat2 & 0.71 & \underline{1.66} & 3.57 & 3.38 & 3.43 & 0.86 \\
    VideoLLaMA 2 & \textbf{\underline{0.78}} & 1.45 & 3.05 & 2.80 & 3.50 & 0.80 \\
    VILA-1.5 & 0.66 & 1.50 & 3.25 & 3.65 & 3.65 & 0.75 \\
    LLaVA-Video & 0.64 & 1.57 & 3.41 & 3.53 & 3.75 & 0.78 \\
    \rowcolor{gray!20} \multicolumn{7}{c}{\textbf{Human}} \\
    Human & \textbf{0.78} & \textbf{1.98} & 3.38 & 3.06 & 3.29 & \textbf{0.94} \\
    \bottomrule
  \end{tabular}
  \caption{\textbf{Results on \taskOne.} The best result for each criterion is shown in bold, and the best model result is underlined.}
  \label{tab:task1}
\end{table}

\subsubsection{Generative}
\noindent\textbf{\taskOne{}--Gen.} Table~\ref{tab:task1} presents the results on \taskOne{}. 
We consider \taskOne{} to be the simplest task in our set, as it does not inherently require abductive reasoning or defeasibility assessment. We observe that models and humans perform within a similar relative margin on both CLIP-Score and LLM-Match metrics. This may be because models are trained on large-scale datasets containing similar event-forecasting tasks. Additionally, these metrics may not fully capture performance in this context, as there are plausible answers at this point, and models are rewarded for almost any relevant prediction they make. 
Human raters also preferred model-generated responses, particularly those from closed-source models like GPT-4o, over the human-written responses. This may be due to style preferences, since model-generated responses are typically grammatical and detailed.


\noindent\textbf{\taskTwo--Gen.} Table~\ref{tab:task2} presents the results for \taskTwo{}. The CLIP metric indicates that closed-source models perform on par with humans, while open-source models lag behind. LLM-Match doesn't provide us with a strong consensus here, as reference-based metrics penalize plausible responses that are different from the references, which makes them unreliable. Human evaluation, thus, provides a more complete perspective. Per human ratings, humans surpass all models by a distinct margin in correctness, thoughtfulness, and level of detail. This corroborates our intuition that \taskTwo{}, requiring abductive reasoning, is significantly harder than \taskOne{}. We observe that VILA-1.5 performs especially well on level of detail, as it tends to generate longer responses, though these are not as correct as human responses.


\setlength{\tabcolsep}{3pt} 
\begin{table}[t]
  \centering
  \begin{tabular}{lcccccc}
    \toprule
    \multirow{2}{*}{\textbf{Model}} & \multicolumn{2}{c}{\textbf{Automatic}} & \multicolumn{4}{c}{\textbf{Human Rating}} \\
    & \textbf{CLIP} & \textbf{LLM-M} & \textbf{C} & \textbf{T} & \textbf{D} & \textbf{V} \\
    \midrule
    \rowcolor{gray!20} \multicolumn{7}{c}{\textbf{Closed Source}} \\
    GPT-4o & \textbf{\underline{0.78}} & 2.08 & 3.41 & \underline{3.28} & 3.53 & \underline{0.78} \\
    Gemini 1.5 Pro & 0.73 & 2.14 & 3.05 & 3.10 & 3.45 & 0.65 \\
    \rowcolor{gray!20} \multicolumn{7}{c}{\textbf{Open Source}} \\
    VideoChat2 & 0.68 & 1.91 & \underline{3.60} & 2.90 & 3.20 & 0.75 \\
    VideoLLaMA 2 & 0.61 & 1.17 & 2.35 & 2.40 & 2.10 & 0.45 \\
    VILA-1.5 & 0.50 & \textbf{\underline{2.18}} & 3.25 & 3.55 & \textbf{\underline{4.00}} & 0.75 \\
    LLaVA-Video & 0.58 & 1.70 & 2.70 & 2.35 & 2.70 & 0.55 \\
    \rowcolor{gray!20} \multicolumn{7}{c}{\textbf{Human}} \\
    Human & 0.77 & 1.92 & \textbf{4.11} & \textbf{3.89} & 3.89 & \textbf{0.79} \\
    \bottomrule
  \end{tabular}
  \caption{\textbf{Results on \taskTwo.} The best result for each criterion is shown in bold, and the best model result is underlined.}
  \label{tab:task2}
\end{table}

\begin{figure*}[ht]
  \centering
  \includegraphics[width=1.0\linewidth]{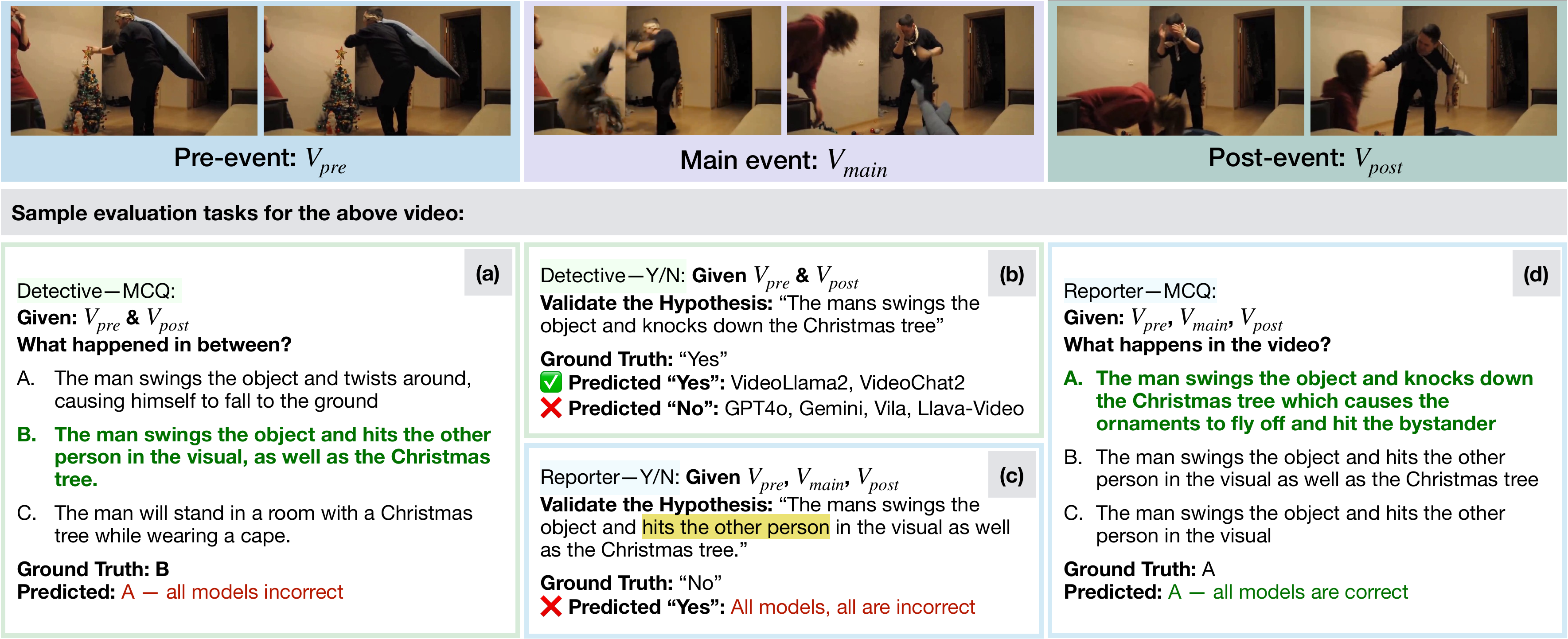}
   \caption{{\bf Qualitative results on MCQ and Y/N variants.} In the video, a man swings a pillow at the Christmas tree, causing ornaments to fly towards the lady. Examples (a), (b), (c) and (d) are task questions from our dataset.}
   \label{fig:quali:descr}
   \vspace{-0.2cm}
\end{figure*}
\setlength{\tabcolsep}{2.7pt} 
\begin{table}[t]
  \centering

  \begin{tabular}{lcccccc}
    \toprule
    \multirow{2}{*}{\textbf{Model}} & \multicolumn{2}{c}{\textbf{LLM-M}} & \multicolumn{4}{c}{\textbf{Human Ratings}} \\
    & \textbf{CLIP} & \textbf{LLM-M} & \textbf{C} & \textbf{T} & \textbf{D} & \textbf{V} \\
    \midrule
    \rowcolor{gray!20} \multicolumn{7}{c}{\textbf{Closed Source}} \\
    GPT-4o & 0.65 & \underline{2.99} & 4.10 & 3.50 & 3.70 & \underline{0.90} \\
    Gemini 1.5 Pro & 0.50 & 2.49 & \underline{4.25} & \textbf{\underline{4.75}} & \textbf{\underline{4.75}} & \underline{0.90} \\
    \rowcolor{gray!20} \multicolumn{7}{c}{\textbf{Open Source}} \\
    VideoChat2 & 0.61 & 2.11 & 3.90 & 3.00 & 3.00 & 0.75 \\
    VideoLLaMA 2 & \underline{0.66} & 2.04 & 2.45 & 2.70 & 3.40 & 0.40 \\
    VILA-1.5 & 0.40 & 2.08 & 3.45 & 3.85 & 3.90 & 0.60 \\
    LLaVA-Video & 0.57 & 2.36 & 3.35 & 3.20 & 3.40 & 0.85 \\
    \rowcolor{gray!20} \multicolumn{7}{c}{\textbf{Human}} \\
    Human & \textbf{0.76} & \textbf{3.23} & \textbf{4.40} & 3.85 & 3.90 & \textbf{0.95} \\
    \bottomrule
  \end{tabular}
  \caption{\textbf{Results on \taskThree.} The best result for each criterion is shown in bold, and the best model result is underlined.}
  \label{tab:task3}
\end{table}

\noindent\textbf{\taskThree--Gen.} The \taskThree--Gen variant resembles a captioning task but has the additional challenge of describing an unexpected event. In this task, we observe that most models default to simple captioning and often lack the specificity needed to describe the actual unexpected event. This trend is evident in Table \ref{tab:task3}, where humans outperform all models on both CLIP and LLM-Match metrics. Human evaluations further reveal that humans are not only slightly more accurate but also more likely to be precise. Notably, Gemini 1.5 Pro scores exceptionally well in depth and detail, due to generating responses that are much longer than human responses, with an average length of 110.75 words compared to 20.78 words by humans.


\subsection{Qualitative Results}
\label{sec:analysis:qualitative}

Figure~\ref{fig:quali:descr} demonstrates the model predictions for the MCQ and Y/N variants for an instance in our dataset. In (a), we show an example from \taskTwo--MCQ where all models failed. This example requires models to distinguish between the two individuals in the video and determine that only the woman (but not the man) falls to the ground. 
In (b), in order to correctly answer the \taskTwo--Y/N question, 
models need to observe that the Christmas tree has been knocked down. 
Both closed-source models failed on this example, while VideoLLaMA 2 and VideoChat2 answered it correctly. 
Part (c) shows the \taskThree--Y/N question, where despite access to the full video, models failed to recognize that the hypothesis is false due to the phrase  ``hits the other person''. Finally, part (d) shows the \taskThree--MCQ question, where all models identified the correct option. 

We leave qualitative results for the generative variant to the Appendix (\ref{app:ResultsQuali}). Our qualitative results reveal that, while our questions are intuitive for humans, they present a complex challenge in visuo-linguistic understanding and reasoning to models.





\section{Analysis}
\label{sec:analysis}
\label{sec:analysis}

We conduct a series of experiments to further study the model capabilities along the aspects of perception, comprehension and reasoning (\S\ref{sec:analysis:perception_comp_reason}), a Chain-of-Thought approach (\S\ref{sec:analysis:cot}), and evaluating models on a hard subset (\S\ref{sec:analysis:hard_set}).

\subsection{Perception, Comprehension and Reasoning}
\label{sec:analysis:perception_comp_reason}

\begin{table}[h!]
  \centering
  \begin{tabular}{lccc}
    \toprule
    \textbf{Model} & \textbf{Baseline (\%)} &  \textbf{+P (\%)} & \textbf{+PC (\%)} \\
    \midrule
    LLaVA-Video & 58.6 & 65.0 \color{green}(+6.4) & 68.6 \color{green}(+10.0)\\
    \bottomrule
  \end{tabular}
  \caption{\textbf{Does adding human perception (P) and comprehension (C) help?} Results on a subset of \taskTwo{} MCQ indicate that the gain in performance is significant.}
  \label{tab:percepcomp}
\end{table}

Answering an abductive reasoning question (\taskTwo{}) requires three key steps: (1) \textbf{perception} of objects, people, and actions in the videos, (2) \textbf{comprehending} the flow of events in the video based on the differences between $V_{pre}$ and $V_{post}$, and (3) abductive \textbf{reasoning} about what could be happening in the middle. We investigate the models' capabilities along each of these aspects by substituting system components with corresponding human-written inputs. In particular, to factor out perception, we include in the prompt the human-written captions for $V_{pre}$ and $V_{post}$ (collected independently), and for comprehension we provide human-written comparisons between $V_{pre}$ and $V_{post}$. See Appendix~\ref{app:Analysis:PC} for the annotation details. 

Evaluating on a subset of 150 MCQ questions with LLaVA-Video (see Table~\ref{tab:percepcomp}), we observe that performance improves by significant margins when perception (+6.4\%) or perception and comprehension (+10\%) are provided. This suggests that current models can improve on foundational perception and comprehension abilities.

\subsection{Chain-of-Thought and Reasoning}
\label{sec:analysis:cot}

\begin{table}[h!]
  \centering
  \begin{tabular}{lcccc}
    \toprule
     & \multicolumn{2}{c}{\taskTwo{}} & \multicolumn{2}{c}{\taskThree{}} \\
     Model & \textbf{Base} &  \textbf{ +CoT} & \textbf{Base} &  \textbf{ +CoT}\\
    \midrule
     LLaVA-Video & 55.6 & 58.0 \color{green}(+2.4) & 69.5 & 68.0 \color{red}(-1.5) \\
    GPT-4o & 71.8 & 77.1 \color{green}(+5.3) & 75.4 & 71.8 \color{red}(-3.6)\\
    \bottomrule
  \end{tabular}
  \caption{\textbf{Does CoT help?} Results on a subset of \taskTwo{} and \taskThree{} MCQ.}
  \label{tab:CoT}
  \vspace{-0.2cm}
\end{table}

Chain-of-thought (CoT) reasoning requires a model to come up with a step-by-step reasoning chain before arriving at a final answer. It is often shown to improve performance in reasoning tasks \cite{wei2023chainofthoughtpromptingelicitsreasoning}. We evaluated the best performing open source and closed source models, LLaVA-Video and GPT-4o with CoT reasoning on a random sample of 150 questions for each of \taskTwo{} and \taskThree{}. We ask models to provide step-by-step reasoning before selecting an MCQ answer. Our results in Table~\ref{tab:CoT} show that CoT improves the performance on \taskTwo{} while worsening the performance on  \taskThree{}. Qualitative analysis show that, on LLaVA-Video, only 14/300 answers actually produced a reason at all. With GPT-4o (Appendix~\ref{app:Analysis:CoTquali}), we see good-quality reasoning steps, but at times, GPT-4o makes assumptions about the outcomes in $V_{post}$, or predisposition to how things behave (e.g. a garbage truck picks up garbage, yet in the video it had malfunctioned), leading to incorrect answers.

\subsection{A Hard Subset}
\label{sec:analysis:hard_set}
\begin{table}[h!]
  \centering
  \begin{tabular}{lcccc}
    \toprule
    \textbf{Model} & \textbf{Base} &  \textbf{Hard} & \textbf{Easy} & $\mathbf{\Delta}$ \\
    \midrule
    GPT-4o  & 65.1 & 57.0 & 67.1 & -10.1 \\
    Gemini-1.5 Pro & 58.7 & 54.3 & 59.8 & -4.7 \\
    VideoChat2 & 29.9 & 26.2 & 30.7 & -4.5 \\
    VILA & 53.5 & 48.3 & 54.8 & -7.8 \\
    VideoLLaMA 2 & 51.8 & 45.5 & 53.3 & -6.5 \\
    LLaVA-Video  & 55.9 & 48.9 & 57.6 & -8.7 \\
    \bottomrule
  \end{tabular}
  \caption{\textbf{Results on the challenging subset of \taskTwo{}--MCQ.} $\mathbf{\Delta}$ is the difference in performance between ``hard'' and ``easy''.}
  \label{tab:mcqhardsubset}
  \vspace{-0.2cm}
\end{table}
Does accuracy vary depending on the predictability of the events? We consider the subset of questions where humans failed to correctly guess what is happening in the video until the entire video was revealed. We identify the hard subset by selecting MCQs for which all the annotations from \taskTwo{} were marked as invalid in \taskThree{}. Table~\ref{tab:mcqhardsubset} shows as much as a 10.1\% drop in performance on the hard subset compared to the easy subset, suggesting models may struggle with highly unpredictable events.

\section{Conclusions}
\label{sec:conclusions}

\bench{} is a novel task to evaluate both abductive and defeasible reasoning with unexpected events. Our benchmark reveals key limitations in VLMs: deficiencies in perception and comprehension, difficulty identifying nuanced information across visual and textual modalities, and challenges in detecting and reasoning about sudden scene changes. Addressing these limitations is crucial step in models that promise to gain innately human capabilities (such as understanding humor), and are perceptually faithful and logical. We hope \bench{} drives progress toward VLMs that truly reason beyond learned correlations.



\section{Acknowledgments}
\label{sec:ack}
This work was funded, in part, by the Vector Institute for AI, Canada CIFAR AI Chair, NSERC CRC, NSERC DG and Accelerator Grants, the Nanyang Associate Professorship, and the National Research Foundation Fellowship (NRF-
NRFF13-2021-0006), Singapore. Hardware resources used in preparing this research were provided, in part, by the Province of Ontario, the Government of Canada through CIFAR, and companies sponsoring the Vector Institute. 


{
    \small
    \bibliographystyle{ieeenat_fullname}
    \bibliography{main}
}

\clearpage
\setcounter{page}{1}
\maketitlesupplementary
\renewcommand{\thesection}{\Alph{section}}
\setcounter{section}{0}


\newenvironment{prompt}
  {\ttfamily\par\setlength{\parindent}{0pt}}
  {\par}

\section{Reasoning Types}
\label{app:reasoningtypes}

\begin{table}[ht]
  \centering
  \footnotesize
  
  \begin{tabular}{lccc}
    \toprule
    \textbf{Task}           & \textbf{Abductive} & \textbf{Defeasible} & \textbf{Commonsense} \\ \hline
    \textbf{\taskOne-Gen}   & \(\times\)         & \(\times\)          & \(\checkmark\)       \\
    \textbf{\taskTwo-Gen}   & \(\checkmark\)     & \(\times\)          & \(\checkmark\)       \\
    \textbf{\taskTwo-MCQ}   & \(\checkmark\)     & \(\checkmark\)        & \(\checkmark\)       \\
    \textbf{\taskTwo-Y/N}   & \(\checkmark\)     & \(\checkmark\)      & \(\checkmark\)           \\
    \textbf{\taskThree-Gen} & \(\times\)         & \(\times\)      & \(\checkmark\)       \\
    \textbf{\taskThree-MCQ} & \(\times\)         & \(\checkmark\)         & \(\checkmark\)       \\
    \textbf{\taskThree-Y/N} & \(\times\)         & \(\checkmark\)      & \(\checkmark\)       \\
    \bottomrule
  \end{tabular}
  \caption{Types of reasoning in BlackSwanSuite.}
  \label{tab:reasoning_tasks}
\end{table}

As shown in Table~\ref{tab:reasoning_tasks}, our benchmark evaluates three fundamental reasoning capabilities through carefully structured tasks. \taskTwo{} assesses \emph{abductive reasoning}, requiring models to infer the most plausible cause of post-events ($V_{post}$) given pre-event contexts ($V_{pre}$).  

Both \taskTwo{} and \taskThree{} incorporate \emph{defeasible reasoning}, presented in two complementary formats: (1) \textbf{Multiple-choice questions (MCQ)}, which assess hypothesis selection through comparative analysis, and (2) \textbf{Yes/No (Y/N) validation}, which requires direct evaluation of a hypothesis when new context emerges. While related, these formats demand distinct reasoning skills—MCQs enable relative comparisons between alternatives, whereas Y/N validation necessitates absolute judgments about specific hypotheses under evolving video contexts.  

To complete the evaluation spectrum, our \emph{generation tasks} (\taskOne-Gen and \taskThree-Gen) assess open-ended prediction of unexpected events, addressing a critical gap in existing video reasoning benchmarks. Collectively, all tasks evaluate \emph{temporal processing} and \emph{visual commonsense}, particularly the ability to interpret and anticipate atypical events.  

With \taskOne{}, \taskTwo{}, and \taskThree{} defined, we next describe our data collection process for constructing generative (\textbf{Gen}), multiple-choice (\textbf{MCQ}), and yes/no validation (\textbf{Y/N}) questions.

\section{Data Collection and Annotations}
\label{app:DataCollection}

\begin{figure}
  \centering
  \includegraphics[width=0.7\linewidth]{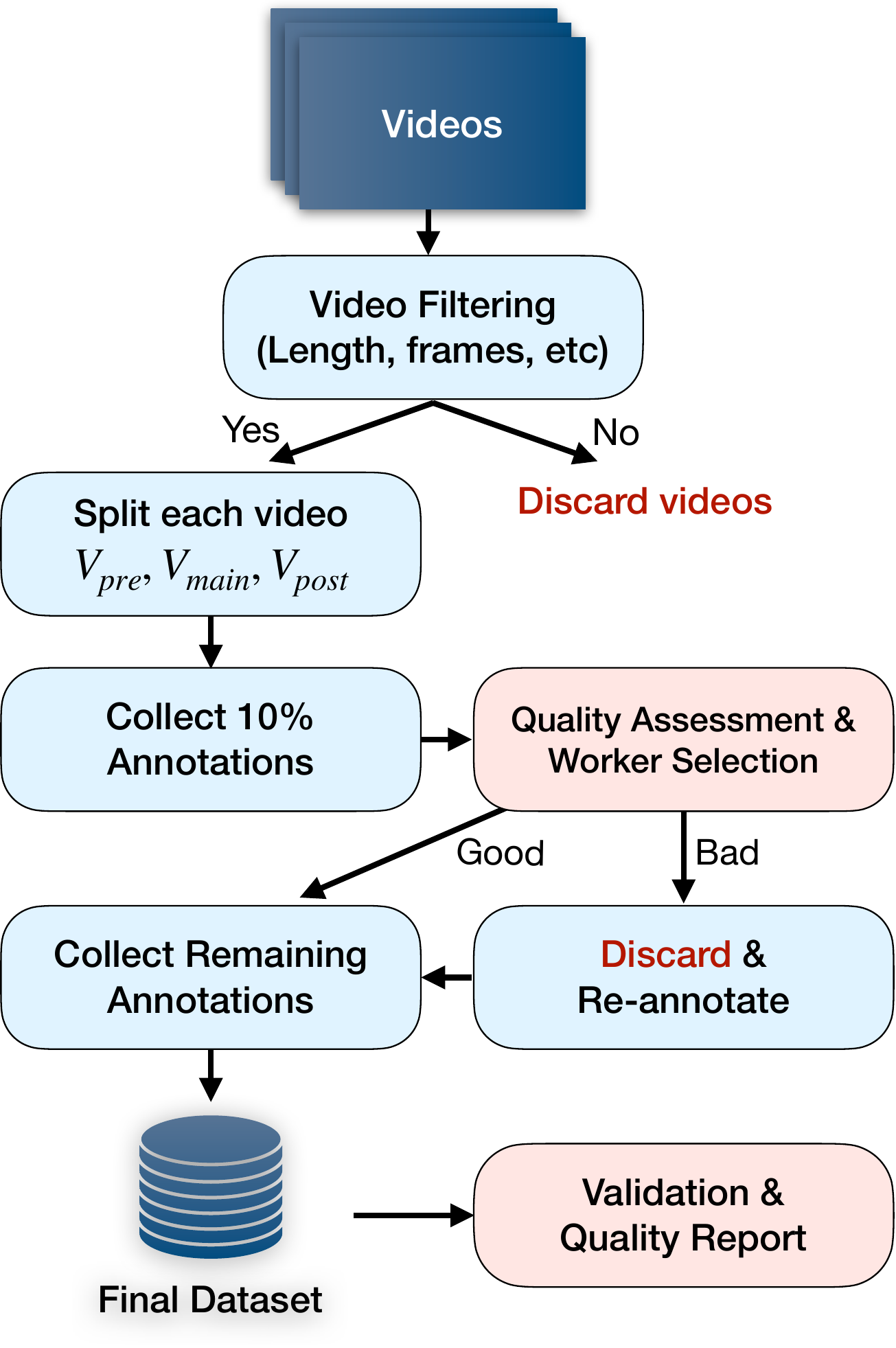}
   \caption{\textbf{Data Collection Process.} We start by filtering videos that adhere to our dataset requirements, such that they can be split into $V_{pre}$, $V_{main}$ and $V_{post}$. With 10\% of data, we collect annotations to select the best annotators. With these annotators, we collect the full dataset, and report dataset quality.}
   \label{fig:datacollection}
\end{figure}

\subsection{Splitting Videos and Collecting Annotations}
\label{app:DataCollectionSplitAlgo}

To automatically split a video $V$ into its parts, $V_{pre}, V_{main},$ and $V_{post}$, we use an automatic scene splitter to clean the video clip, and use heuristics described in Algorithm~\ref{alg:split} to perform the cut, based on the main event time, $t$. Note that we acquire both the video $V$ and the main event time $t$ from the Oops! dataset \cite{epstein2020oops}. 

First, to ensure that the video $V$ does not contain multiple different scenes, we use PySceneDetect's AdaptiveDetector\footnote{\url{https://www.scenedetect.com/docs/latest/api/detectors.html\#scenedetect.detectors.adaptive_detector.AdaptiveDetector}} with window width set to 5 frames to identify the scenes in the video. In case 4 or more scenes are found, we discard the video. If 2 or 3 scenes are found, and if the scene change is in the beginning and/or the end of the video, we trim the beginning and/or the end. Moreover, a change that coincides with the event time $t$ can occur when major changes to the scene happen as a part of the surprising event, \eg a light bulb going off, making the entire scene dark, is allowed. Following these steps, we get a cleaned-up and trimmed video $V^*$.

\newcommand{\CommentLine}[1]{
    \State // \textit{#1}
}

\begin{algorithm}
\caption{Video splitting}\label{alg:split}
\begin{algorithmic}[1]
\Require video $V^*$, main event time $t$ 
\Ensure the three parts of the video $V_{pre}, V_{main}, V_{post}$ 




\CommentLine{Trim the start and end of the video, in case there are remnants of adjecent scenes.}
\State $V^* \leftarrow \operatorname{trim}(V^*, 0.17~\text{sec}, \text{start})$ 
\State $V^* \leftarrow \operatorname{trim}(V^*, 0.17~\text{sec}, \text{end})$ 

\CommentLine{Cut the video into parts}
\State $V_{pre} \leftarrow [0, 0.8 \cdot t]$
\State $V_{main} \leftarrow [0.8 \cdot t, 0.8 \cdot \operatorname{length}(V^*)]$ 
\State $V_{post} \leftarrow [0.8 \cdot \operatorname{length}(V^*), \operatorname{length}(V^*)]$

\CommentLine{Discard videos that are shorter than 1 second}
\If{$\operatorname{any}(\operatorname{length}(v) < 1 \text{~sec~} \forall v \in \{ V_{pre}, V_{main}, V_{post} \}) $} 
    \State \Return Null
\EndIf

\State \Return $V_{pre}, V_{main}, V_{post}$

\end{algorithmic}
\end{algorithm}

The algorithm receives the cleaned-up video and the average event time based on the annotations. To prevent parts of the previous or next video clip to be visible (due to the window width set to 5 frames), we trim the ends of the video by 0.17 seconds (lines 1-3). We then cut $V^*$ into its three parts based on the mean event localization time $t$, such that $V_{pre}$ ends shortly before $t$, $V_{post}$ is the last 20\% of the video, and $V_{main}$ shows the main event (lines 4-7). Finally, we discard any videos for which at least one of the 3 parts is less than 1 second long (lines 8-11). This ensures that there is enough content in each part of the event. Through empirical analysis, we found that this method generally yields $V_{pre}, V_{main}, V_{post}$ where $V_{pre}$ doesn't reveal exactly what unexpected event is about to happen, but allows for reasonable guesses (ideal for defeasible reasoning), and $V_{post}$ only shows the outcome of the event, allowing for abductive reasoning.

Finally, we collect annotations using the template presented in Figure \ref{fig:mainannottemplate}. We have the following conditions to participate as an annotator for our task: 

\begin{itemize}
    \item Location: Ireland, Canada, New Zealand, United Kingdom, United States, Australia
    \item Age: 18-65 year old
    \item Education: Bachelor's degree (for example: BA, AB, BS), Master's degree (for example: MA, MS, MEng, MEd, MSW, MBA), Professional degree (for example: MD, DDS, DVM, LLB, JD), Doctorate degree (for example: PhD, EdD)
\end{itemize}

In each step of the process, annotators must write at least 8 words, and all answers are required and cannot be skipped. 

\subsection{Data Quality Validation}
\label{app:DataCollectionQValid}

\begin{figure*}
  \centering
  \includegraphics[width=0.9\linewidth]{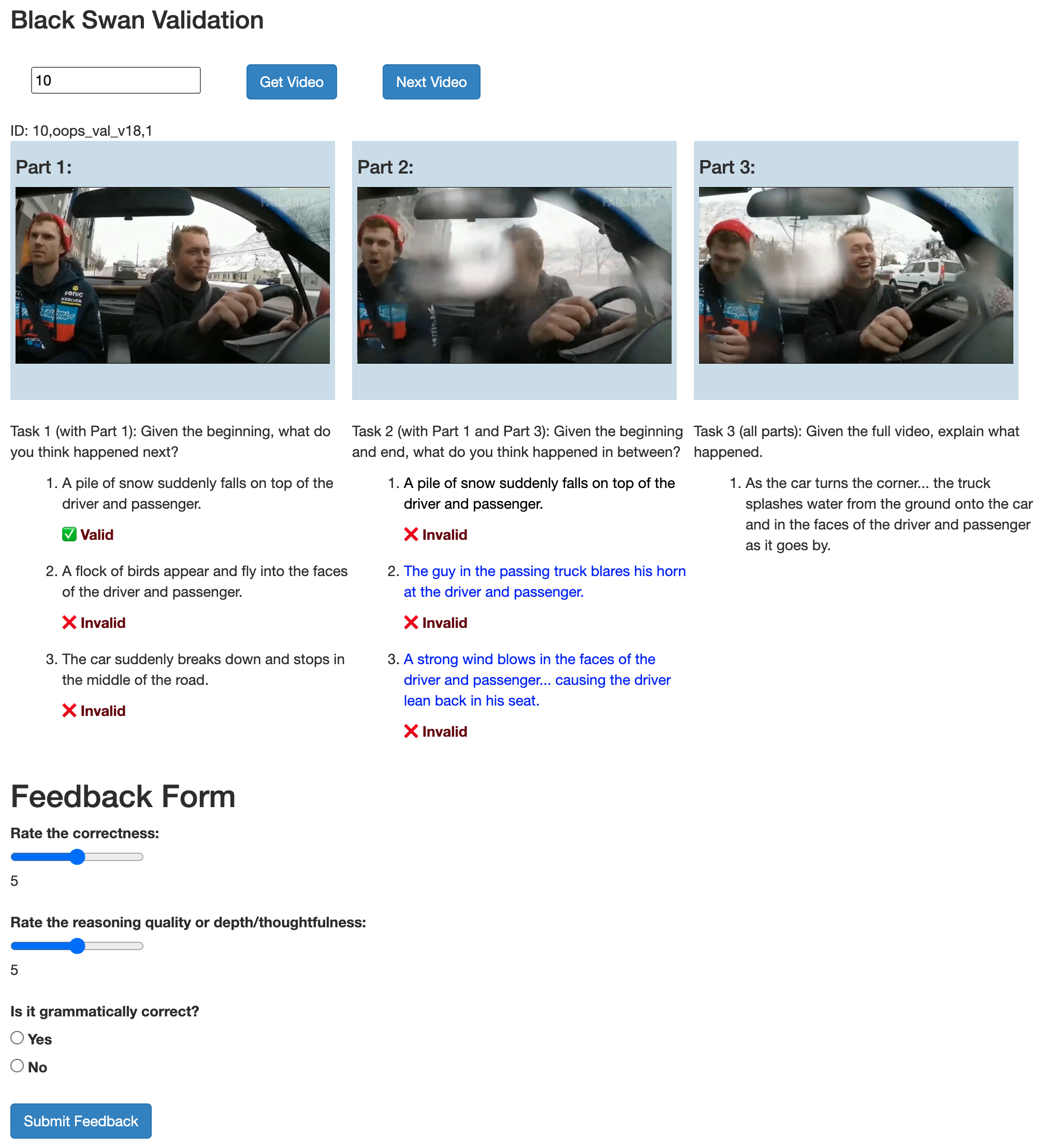}
   \caption{\textbf{Template for Dataset Validation.}}
   \label{fig:datasetval}
\end{figure*}

We ask two students (experts) from the lab, who did not contribute to this project in any other way, to independently verify 60 randomly sampled annotation instances from our dataset. Experts are first explained all the tasks, and are provided will all parts of the video along with annotations from all three steps. We provide them with the UI in Fig. \ref{fig:datasetval}, where they can grade each annotation on the basis of correctness (where one mistake can indicate a deduction of one point), level of detail (do the descriptions have sufficient detail in order to easily discern people/objects and actions?), and grammar (are the descriptions reasonably well written, it does not have to be perfect). We report the average scores between the two individuals.

Furthermore, following this process, we ask the experts for feedback. We received the following feedback:
\begin{itemize}
    \item Some annotations may contain minor mistakes, where the annotator did not look at the video carefully, and therefore marked a description that could have been valid as invalid (or vice versa). A common cause of this was the low video quality. For example, an annotator invalidated a claim that a car hit the sidewalk, though in $V_{post}$, the car does hit the sidewalk, but the low quality of the $V_{post}$ made difficult to observe.
    \item In some cases, annotators may have missed obvious explanations of what is going on. In such cases, their descriptions may be reasonably correct, but not a true description of what is happening.
    \item Regarding level of detail, the experts suggested that in some cases, the sentences were too short and did not describe the scene sufficiently.
    \item Grammatically speaking, the only times it was marked as "no" is when there were multiple grammatical errors across multiple descriptions for the same video.
\end{itemize}

\subsection{Annotator Statistics}
\label{app:annotatorstats}

To evaluate the diversity in the annotators of our dataset, we consider age, education level, gender and country of the annotators. Here are the statistics: 

\paragraph{Age:}
\begin{itemize}
    \item Average Age: \textbf{37.5}
    \item Median Age: \textbf{36}
    \item Minimum Age: \textbf{22}
    \item Maximum Age: \textbf{61}
\end{itemize}

\paragraph{Education Level:}
\begin{itemize}
    \item Bachelor's degree (for example: BA, AB, BS): \textbf{60.3\%}
    \item Master's degree (for example: MA, MS, MEng, MEd, MSW, MBA): \textbf{37.76\%}
    \item Doctorate degree (for example: PhD, EdD): \textbf{1.94\%}
\end{itemize}

\paragraph{Gender:}
\begin{itemize}
    \item Man: \textbf{48.95\%}
    \item Woman: \textbf{39.5\%}
    \item Not Known: \textbf{11.55\%}
\end{itemize}

\paragraph{Country:}
\begin{itemize}
    \item USA: \textbf{73.85\%}
    \item Canada: \textbf{13.05\%}
    \item New Zealand: \textbf{11.5\%}
    \item UK: \textbf{1.6\%}
\end{itemize}

\section{Metrics}
\label{app:Metrics}

\paragraph{CLIP Score.} We use \texttt{clip-vit-large-patch14} to embed each sentence, and use cosine similarity as the distance measure for any pair of sentences.

\paragraph{LLM-Based Metric} We use Llama 3.1 8B (Huggingface: \texttt{Llama-3.1-8B-Instruct}). We chose to take the mean instead of the max of each pairwise score while comparing the ground truth set. This is because the sparsity of scores (1, 2, 3, 4 or 5) makes it very easy the LLM to rate a max score of 3 or 4 for any pair of sentence sets, yielding results that show very little distinction between different models. Instead, taking the mean allows us to also measure the diversity of generations, which indicates how aligned LLMs are with humans for all their predictions, across all samples generated. 

The prompt for LLM-Match is:

\begin{prompt}
You are an AI assistant tasked with evaluating how well a given response aligns with the provided ground truth. Focus on the semantic similarity between the two texts. Your assessment should produce a single integer score between 1 and 5:

5: The response matches the ground truth perfectly.

1: The response is entirely different from the ground truth.

Please return your evaluation in the following format:

Reason: A brief, ten-word explanation for your score.

Score: Your score.

Ground Truth:    
{ground\_truth}

Response to Score:
{model\_generated}
    
\end{prompt}

\paragraph{BLEU and ROUGE.} Tasks in \bench{} often involve generating hypotheses, explanations, or descriptions, which are inherently open-ended. Multiple valid answers may exist, and these may differ significantly in phrasing from the reference answer. N-gram based metrics such as BLEU and ROUGE penalize such variations, despite them being semantically correct. Hence, these metrics may not accurately measure correctness of generations. However for the sake of completeness, we report them in Table ~\ref{tab:bleu_rouge}.
\setlength{\tabcolsep}{1pt}
\begin{table}[ht]
\centering
\footnotesize
\begin{tabular}{lcccccc}
\toprule
\textbf{Model} & \multicolumn{2}{c}{\textbf{\taskOne}} & \multicolumn{2}{c}{\textbf{\taskTwo}} & \multicolumn{2}{c}{\textbf{\taskThree}} \\
\cmidrule(lr){2-3} \cmidrule(lr){4-5} \cmidrule(lr){6-7}
& \textbf{BLEU} & \textbf{ROUGE-L} & \textbf{BLEU} & \textbf{ROUGE-L} & \textbf{BLEU} & \textbf{ROUGE-L} \\
\midrule
GPT-4o & 0.70 & 27.10 & 0.80 & \textbf{26.06} & \textbf{0.90} & \textbf{23.40} \\
Gemini 1.5 Pro & \textbf{1.26} & \textbf{28.03} & \textbf{1.12} & 24.14 & 0.43 & 11.90 \\
VideoChat 2 & 0.49 & 25.80 & 1.00 & 26.04 & 0.70 & 20.90 \\
VideoLLaMA 2 & 0.67 & 25.20 & 0.11 & 15.42 & 0.52 & 17.30 \\
VILA & 0.80 & 18.77 & 0.83 & 16.59 & 0.60 & 15.30 \\
LLaVA-Video & 0.49 & 20.50 & 0.61 & 24.28 & 0.81 & 20.50 \\
Human & 1.18 & 25.41 & 0.98 & 24.74 & 0.00 & 0.45 \\
\bottomrule
\end{tabular}
\caption{\textbf{BLEU and ROUGE-L results for \taskOne, \taskTwo, and \taskThree{} Gen varaints (\( \times 100 \)).}}
\label{tab:bleu_rouge}
\end{table}

\section{Human Evaluation Setup}
\label{app:HumanEval}

In order to conduct human evaluation for the generative variants, we use the template in Figure \ref{fig:humanevalgen}. We define the four criterion as:
\begin{enumerate}
    \item Correctness (5 Point Scale) Check whether the answer correctly describes what could have happened in part 2 based on part 1.
    \item Depth and Thoughtfulness (5 Point Scale) How much depth in reasoning is covered in the description? Is there sufficient reasoning to support the description?
    \item Level of Detail (5 Point Scale) Does the description have sufficient level of detail to easily identify and differentiate between people/objects and actions?
    \item Visual Contradiction (Yes/No) Does the description follow the content in the video, or contradict the contents of the video? A contradiction happens when the description is impossible, given the video.
\end{enumerate}

We use the same location, age and education filters as used above while collecting annotations. Annotators are always required to write reasons for their choices; accordingly, after running this evaluation for one model for 50 videos, we manually filter out evaluators who have done this task diligently. This gave us 32 annotators who were capable of doing this task well. We use these annotators on all evaluations of model and human written explanations (in the case of \taskOne{}--Gen and \taskTwo{}--Gen, we sample a single explanation from the three generated by the model). 

While we would have liked to conduct this evaluation with multiple annotators for each video, and over a larger set of videos, we believed that this would be very expensive and not a practical use of funds. We paid evaluators \$0.45 per evaluation, as each video and explanation can require 4-5 minutes to evaluate (since reasons must be written). Say we had used 50 videos, and evaluated a single explanation for each task, from all models, 3 separate times by different annotators. At this rate, for 7 models (incl. human), 50 videos, and 3 tasks and 3 annotators, and a 25\% service fee, this would be over \$1770. We instead sought out to compare results qualitatively, which we do in Appendix \ref{app:ResultsQuali}.

\begin{figure*}[h]
  \centering
  \includegraphics[width=1.0\linewidth]{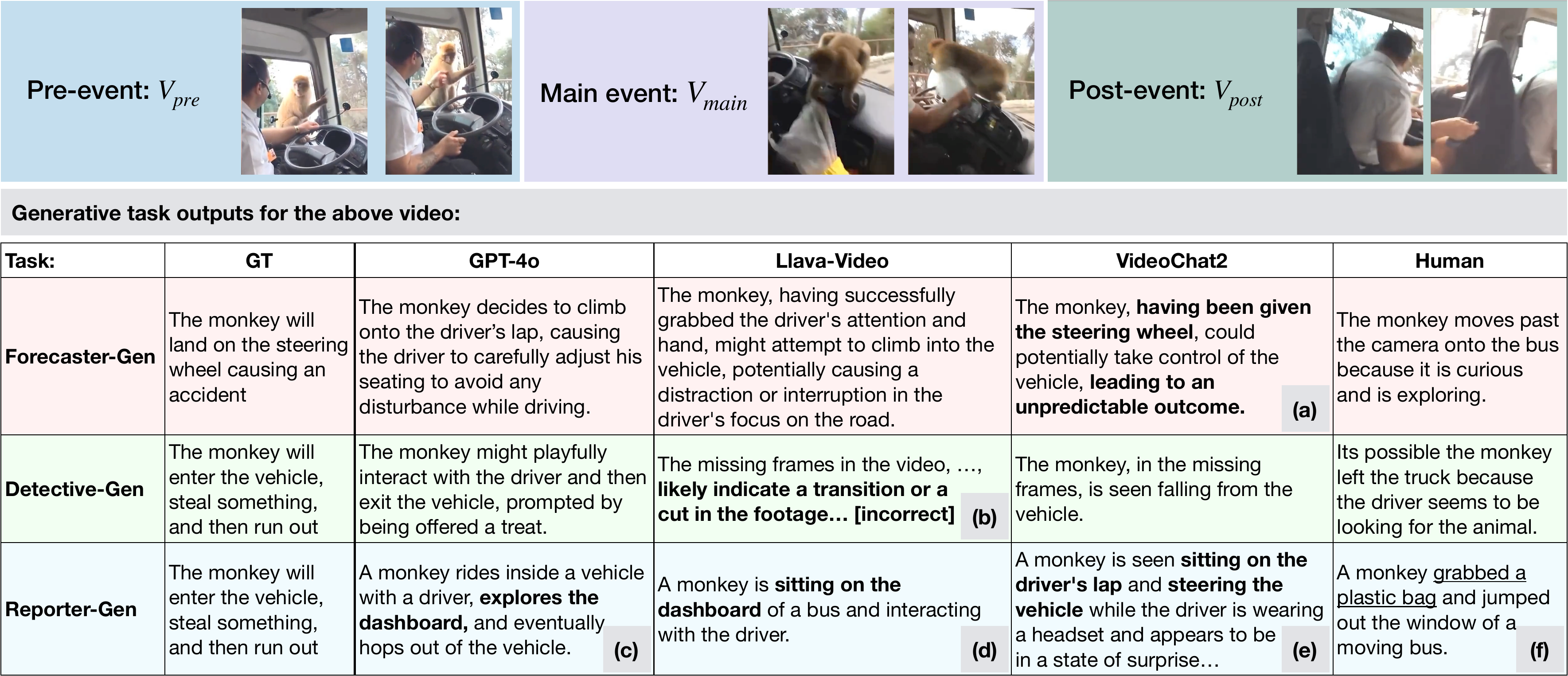}
   \caption{{\bf Qualitative results on Gen variants.} Due to space constrains, only one sample from each model is shown.}
   \label{fig:quali:gen}
\end{figure*}
\begin{figure*} 
    \centering

    \begin{subfigure}[b]{1\textwidth} 
        \centering
        \includegraphics[width=\textwidth, page=3]{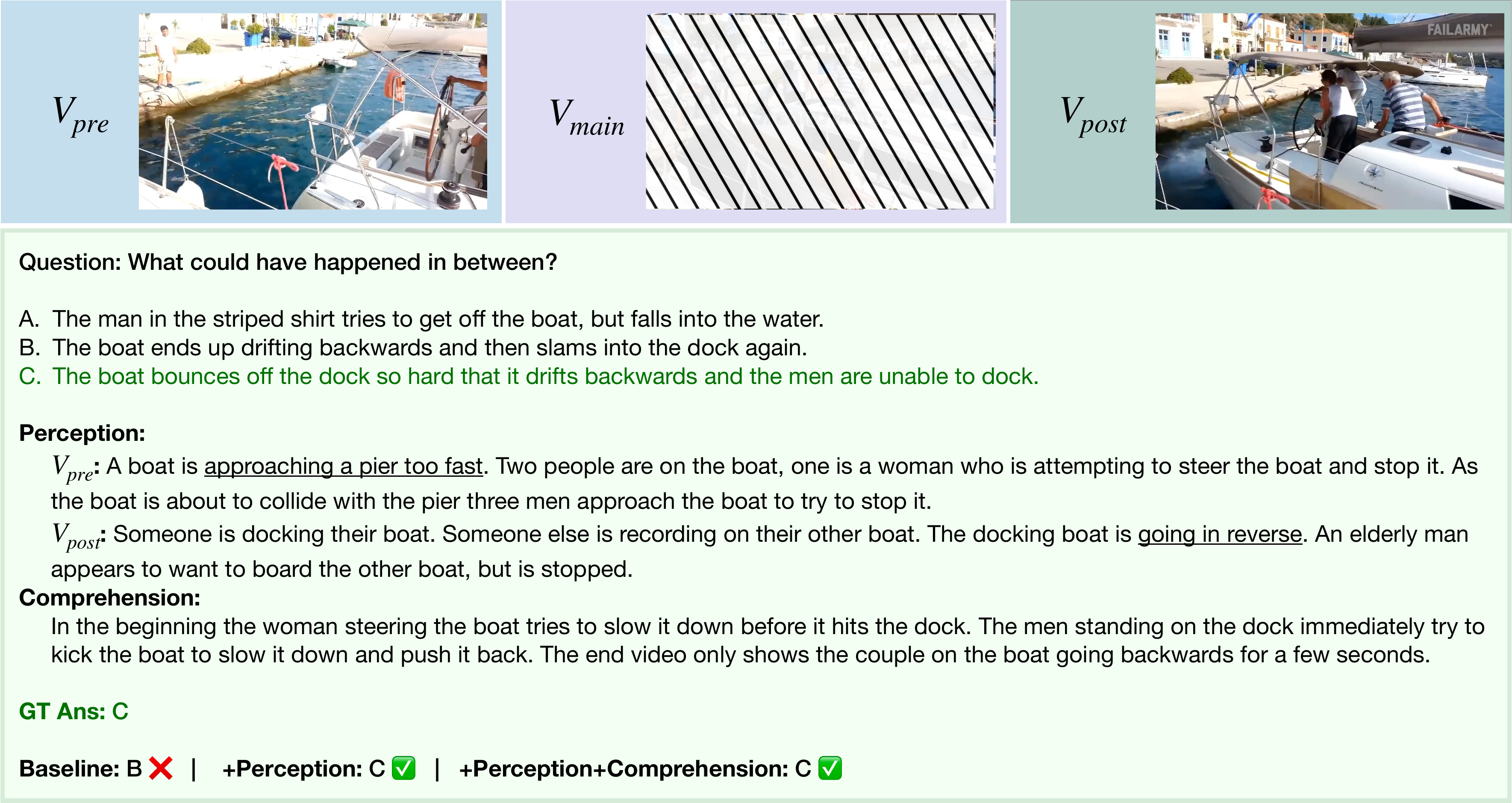}
        \caption{Example where LLaVA-Video benefits from Perception}
    \end{subfigure}
    \vskip\baselineskip

    \begin{subfigure}[b]{1\textwidth}
        \centering
        \includegraphics[width=\textwidth, page=4]{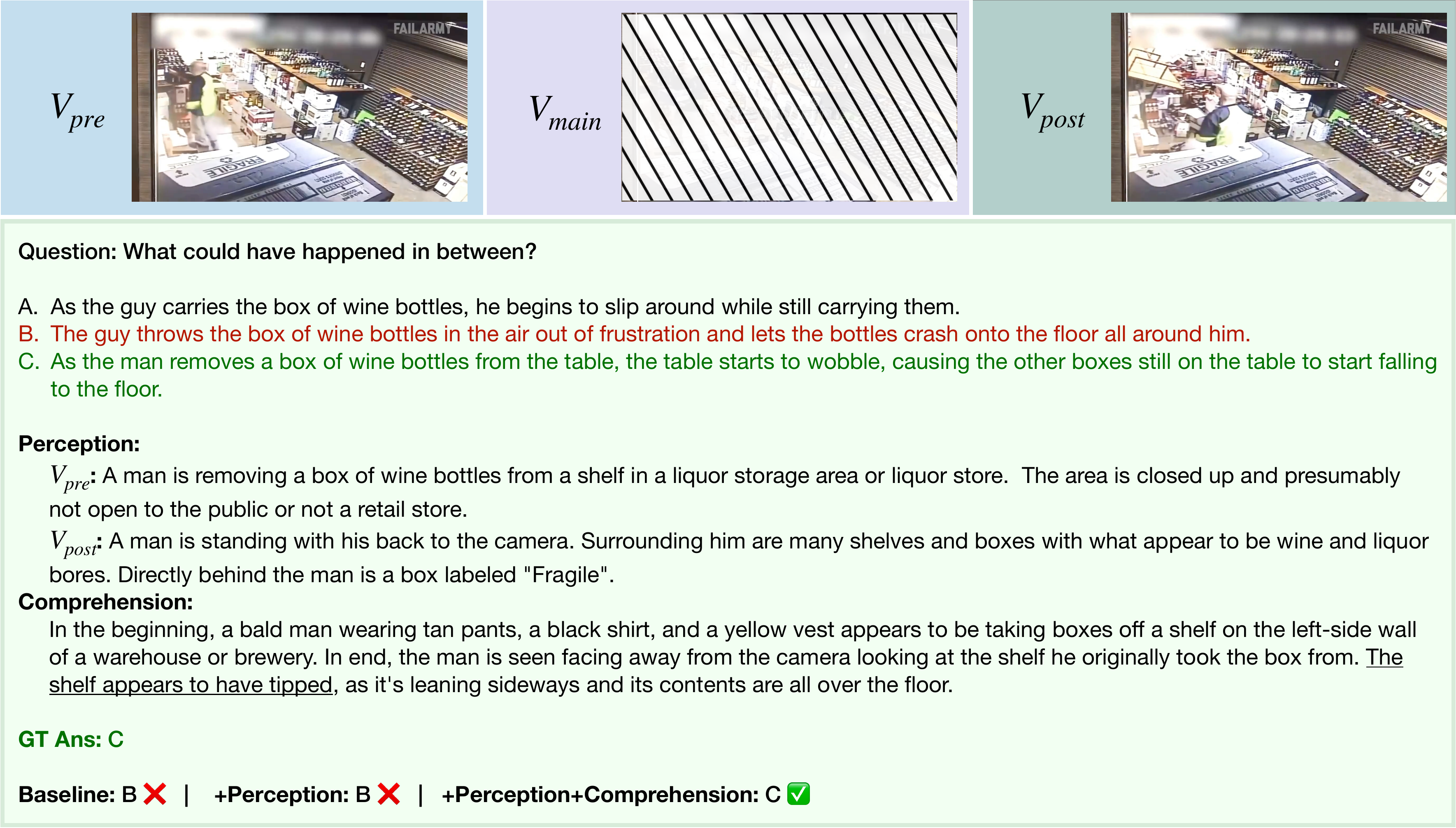}
        \caption{Example where LLaVA-Video benefits from Comprehension}
    \end{subfigure}
    \vskip\baselineskip


    \caption{\textbf{Perception and Comprehension with LLaVA-Video.} (a) shows a case where human-written Perception helps. (b) shows a case where both human-written Comprehension helps.}
    \label{fig:pcegs}
\end{figure*}

\section{Baselines}
\label{app:Baselines}

We test several latest VLMs. Closed-source VLMs include:

\noindent\textbf{1. OpenAI GPT-4o \cite{openai2024gpt4o}} GPT-4o was an important choice for our video-based task because, apart from leading all kinds of VL benchmarks, it is OpenAI's first model that is capable of accepting up to 50 pictures as input in the prompt. We use the GPT-4o model using the OpenAI Batch API\footnote{\url{https://platform.openai.com/docs/guides/batch/overview}}. The model was accessed between October 20th and November 14th, 2024. We feed it 10 uniformly sampled frames for every part of the video.

\noindent\textbf{2. Google Gemini 1.5 Pro \cite{team2024gemini}} We chose to use Gemini 1.5 Pro as it is the most capable VL model that can natively accept video input. This model was accessed between October 20th and November 14th, 2024. When asking questions, we directly upload the entire video clip. In the case of \taskTwo{}, the clip has the $V_{main}$ blacked out.

Open source models include:

\noindent\textbf{3. VideoChat2 \cite{li2024mvbenchvideochat2}} VideoChat2 showed stronger detail and contextual understanding than models prior to it, and even beat GPT-4V at the time of its release on multiple VL tasks. We evaluate the latest VideoChat2 HD (Huggingface: \texttt{videochat2\_hd\_mistral\_7b\_stage4}) model. When asking questions, we directly upload the entire video clip. In the case of \taskTwo{}, the clip has the $V_{main}$ blacked out. Default settings of 16 frames per video input are used, with resolution set to 224.

\noindent\textbf{4. VideoLLaMA 2 \cite{cheng2024videollama2}} VideoLLaMA 2 is capable of understanding both visual and audio signals. Though audio is not a requirement for this task, there may be cases where it can reveal important information about the events in the video. We use the \texttt{VideoLLaMA2.1-7B-16F} model. Again, when asking questions, we directly upload the entire video clip. In the case of \taskTwo{}, the clip has the $V_{main}$ blacked out. Default settings for resolution are used, and max number of frames are 32.

\noindent\textbf{5. VILA \cite{lin2024vila}} VILA 1.5 is the latest in the VILA series of models, one of the first models to support multi-image understanding. We use the \texttt{Llama-3-VILA1.5-8b-Fix} model. We uniformly sample 5 frames for each part of the video.

\noindent\textbf{6. LLaVA-Video \cite{zhang2024llavavideo}} LLaVA was first introduced as a multimodal model with performance rivaling GPT-4. Following the release of LLaVA 1.5 and LLaVA-NeXT, LLaVA-Video is the most advanced version of this model, specifically trained on a new 178K video dataset. Specifically, we use the \texttt{LLaVA-Video-7B-Qwen2} model specification. When asking questions, we directly upload the entire video clip. In the case of \taskTwo{}, the clip has the $V_{main}$ blacked out. 32 frames are used. We also test the 72B LLaVA-Video (\texttt{LLaVA-Video-72B-Qwen2}) model; however, due to compute limitations and setup issues, we were only able to run that model on a subset of the data. The results are detailed in Appendix \ref{app:Results:llava72b}.

Our implementation of all open-source models follows their respective instructions on GitHub and Huggingface, and we tried our best to recreate the same environment as the original developers for each model. We do not do batch inference on any of these models, and instead query them iteratively in chat mode only (each chat conversations for every question is independent). For the generative variant, in order to obtain different results each time, we turn sampling on.

We use the following prompts for each task:

\paragraph{\taskOne--Gen}

\begin{prompt}
    Describe what could happen next, by explaining the sequence of actions leading to the outcome.
\end{prompt}

\paragraph{\taskTwo--Gen}

\begin{prompt}
    What happened in the missing frames (in black) of the video?
\end{prompt}

\paragraph{\taskTwo--MCQ}

\begin{prompt}
    Select the description that indicates what happened in the hidden (black) frames of the video: A. <Option A> B. <Option B> C. <Option C>
\end{prompt}

\paragraph{\taskTwo--Y/N}

\begin{prompt}
    Hypothesis: <hypo>
    
    Given the video clip, does this hypothesis hold? Answer yes or no.
\end{prompt}

\paragraph{\taskThree--Gen}

\begin{prompt}
    Explain what is happening in the video.
\end{prompt}

\paragraph{\taskThree--MCQ}

\begin{prompt}
    Select the description that correctly explains what happens in this video: A. <Option A> B. <Option B> C. <Option C>
\end{prompt}

\paragraph{\taskThree--Y/N}

\begin{prompt}
    Hypothesis: <hypo>
    
    Given the video clip, does this hypothesis hold? Answer yes or no.
\end{prompt}

Depending on the model, we may vary the prompt in small ways. For example, we may use the default system instruction for each model, or ask models to return an answer in only one sentence. In the case of multi-frame models, when we feed frames from the beginning and the end, we specify it as such:
\begin{prompt}
Here is the beginning of the video: <image tokens for V\_pre>

Here is the end of the video: <image tokens for V\_post>

<Question...>
\end{prompt}

\section{Results}
\label{app:Results}

\subsection{Human Baseline}
\label{app:ResultsHumanEval}

\begin{figure*}
  \centering
  \includegraphics[width=0.9\linewidth, page=12]{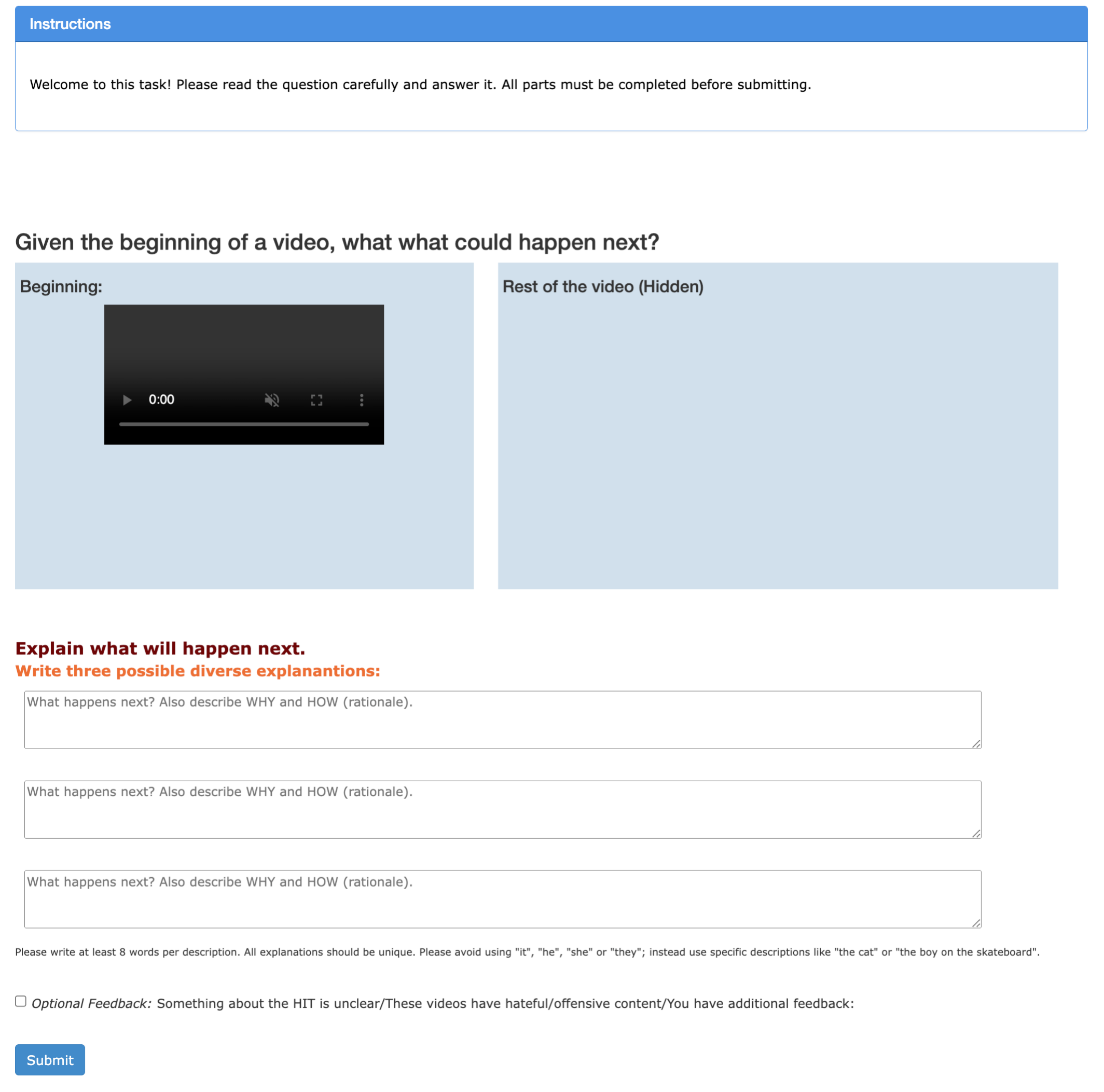}
   \caption{\textbf{Sample template for Human Baseline for the generative variant.}}
   \label{fig:baselinegen}
\end{figure*}

\begin{figure*}
  \centering
  \includegraphics[width=0.9\linewidth, page=10]{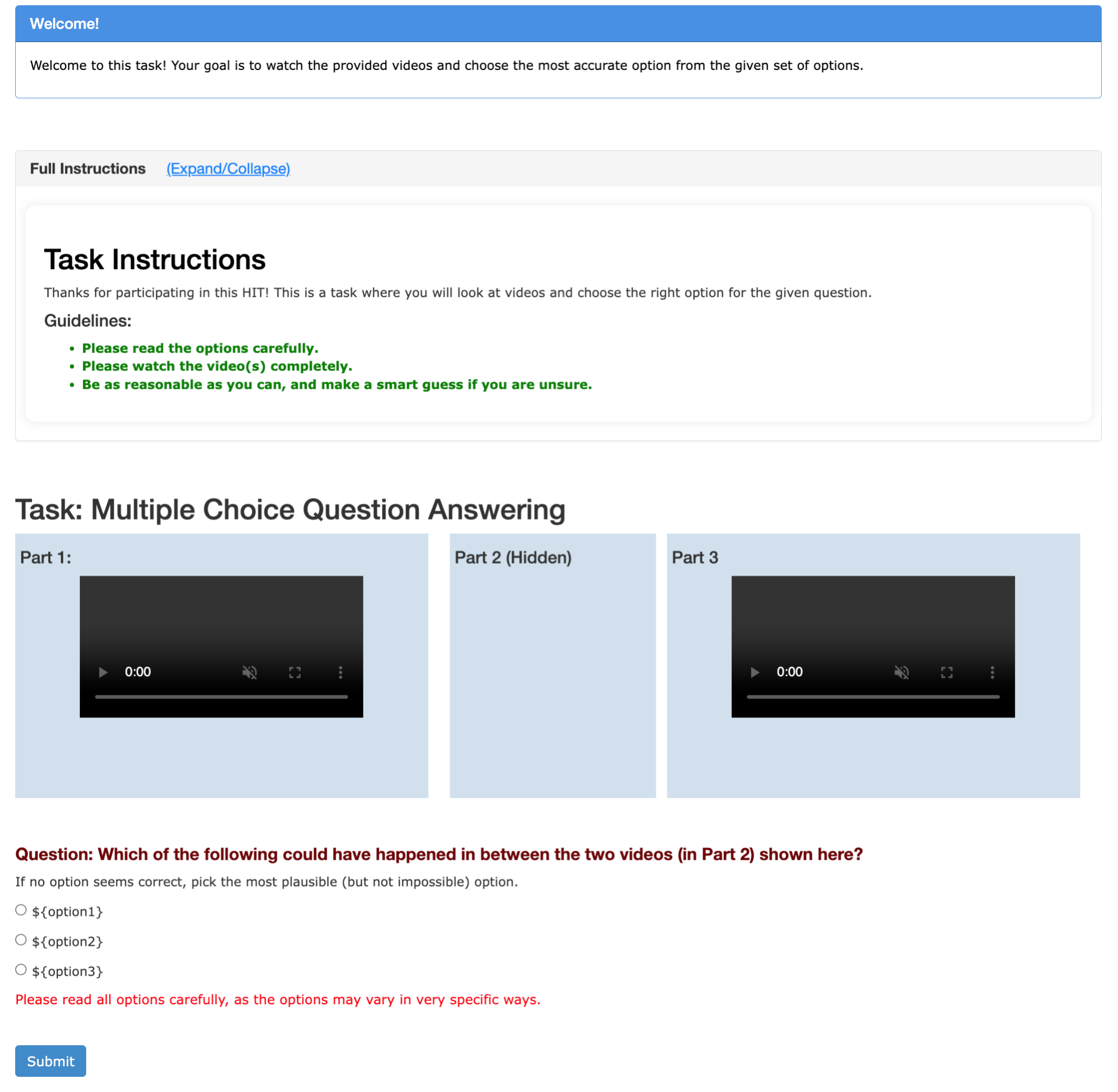}
   \caption{\textbf{Sample template for Human Baseline for the MCQ variant.}}
   \label{fig:baselinemcq}
\end{figure*}

\begin{figure*}
  \centering
  \includegraphics[width=0.9\linewidth, page=11]{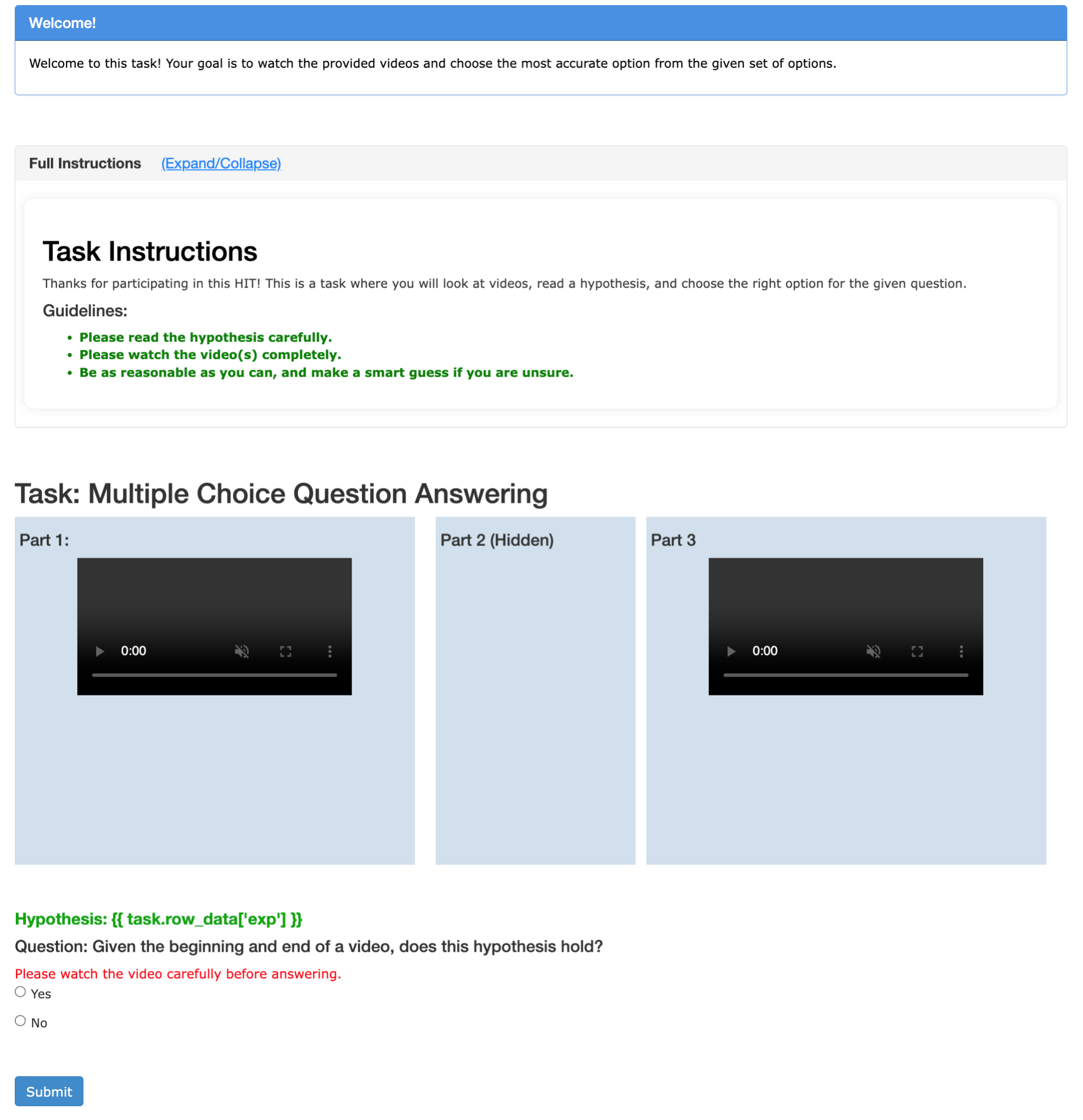}
   \caption{\textbf{Sample template for Human Baseline for the Y/N variant.}}
   \label{fig:baselineyn}
\end{figure*}

We compare models to human performance. We ask humans to do the same tasks as models. For the generative variant, we use the template in Figure \ref{fig:baselinegen} to collect annotations (we build similar templates for other generative variants, with more parts of the videos shown). For the MCQ variant, an example template is shown in Figure \ref{fig:baselinemcq}. Similarly, for the Y/N task, an example template is shown in Figure \ref{fig:baselineyn}.

For each of the MCQ and Y/N variants of \taskTwo{} and \taskThree, we ask two students from the lab to answer 100 randomly sampled questions for MCQ and 150 randomly sampled questions for Y/N and report the maximum score achieved across the two annotators. We do this since we want to measure the maximum achievable score by a human expert on a given task, as an upper bound. For the generative variant of all three tasks, we crowd source new annotations by a single annotator (same CloudResearch setup as before), and compute metrics on the obtained annotations.

\subsection{Additional Qualitative Results}
\label{app:ResultsQuali}

Figure~\ref{fig:quali:gen} demonstrates the outputs for the generative task variants. Models are reasonably good at estimating what could happen next (\taskOne--Gen), albeit with some uncertainty. For example, in (a), VideoChat2 hallucinates the action of having been given the steering wheel, and expresses uncertainty (``leading to an unpredictable outcome'') Having uncertainty, by itself, is technically not wrong, but it is often a way for the model to avoid the question. Uncertainty is better than hallucination or returning an entirely wrong answer. In part (b) (\taskTwo-Gen), LLaVA-Video misunderstands the instruction to hypothesize about what is happening in the missing frames and instead describes why the missing frames are not present. In \taskThree{}--Gen (parts c--e), all models failed to capture the instantaneous event where the monkey picks up the plastic bag and leaves, which lasts less than 2 seconds. The most clear response, in this case, is from the human, who clearly stated that the monkey grabbed the plastic bag (f).

For additional examples, please watch the video \texttt{examples.mp4} in the supplementary zip file.

\subsection{Additional Quantitative Results}
\label{app:Results:llava72b}
\setlength{\tabcolsep}{3pt} 
\begin{table}[ht]
\centering
\begin{tabular}{lcccc}
\toprule
\multicolumn{1}{c}{\multirow{2}{*}{\textbf{Model}}} & \multicolumn{2}{c}{\textbf{\taskTwo}} & \multicolumn{2}{c}{\textbf{\taskThree}} \\
\multicolumn{1}{c}{} & \textbf{MCQ} & \textbf{Y/N} & \textbf{MCQ} & \textbf{Y/N} \\
\midrule
\rowcolor{gray!20} \multicolumn{5}{c}{\textbf{Open Source}} \\
LLaVA-Video-7B & 55.9 & 59.3 & 69.6 & 55.1 \\
LLaVA-Video-72B & 59.91 & 56.93 & 74.91 & 54.47 \\
\rowcolor{gray!20} \multicolumn{5}{c}{\textbf{Human}} \\
Human & \textbf{90.0} & \textbf{85.3} & \textbf{95.3} & \textbf{92.0} \\
\bottomrule
\end{tabular}
\caption{\textbf{Results on MCQ and Y/N variants of \taskOne and \taskTwo} on 72B variant of LLaVA-Video on 20\% of the data.}
\label{tab:70b}
\end{table}

\setlength{\tabcolsep}{2pt}
\begin{table}[ht]
  \centering
  \small 
  \begin{tabular}{lcccccc}
    \toprule
    \multicolumn{1}{c}{\textbf{Model}} & \multicolumn{2}{c}{\textbf{\taskOne}} & \multicolumn{2}{c}{\textbf{\taskTwo}} & \multicolumn{2}{c}{\textbf{\taskThree}} \\
    \cmidrule(lr){2-3} \cmidrule(lr){4-5} \cmidrule(lr){6-7}
    & \textbf{CLIP} & \textbf{LLM-M} & \textbf{CLIP} & \textbf{LLM-M} & \textbf{CLIP} & \textbf{LLM-M} \\
    \midrule
    \rowcolor{gray!20} \multicolumn{7}{c}{\textbf{Open Source}} \\
    LLaVA-V-7B & 0.64 & 1.57 & 0.58 & \underline{1.70} & 0.57 & \underline{2.36} \\
      LLaVA-V-72B & \underline{0.69} & \underline{1.67} & \underline{0.60} & 1.68 & \underline{0.59} & 2.15 \\
    \rowcolor{gray!20} \multicolumn{7}{c}{\textbf{Human}} \\
    Human & \textbf{0.78} & \textbf{1.98} & \textbf{0.77} & \textbf{1.92} & \textbf{0.76} & \textbf{3.23} \\
    \bottomrule
  \end{tabular}
  \caption{\textbf{Results on Gen variants with 72B version on 10\% of data.} Metrics are grouped by tasks, each with CLIP and LLM-M scores.}
  \label{tab:genllava72B}
\end{table}

In Table~\ref{tab:70b} we include the results on a subset of 20\% randomly sampled MCQ and Y/N questions on the 72B LLaVA-Video model, and compare it against the 7B model. We observe that the 72B version significantly outperforms the 7B variant and reaches an accuracy on both MCQ and Y/N close to the best-performing closed-sourced models shown in Table ~\ref{tab:disc}. 

Table~\ref{tab:genllava72B} shows CLIP Score and LLM-Match on the three tasks for the 72B variant, on 10\% of the data. We only generate a single explanation for each task (in general, we generate three responses for \taskOne{} and \taskTwo{} for all other models). We observe that the 72B version performs very similarly to the 7B version.

\section{Leaderboard Results}
\label{app:leaderboard}

\setlength{\tabcolsep}{3pt} 
\begin{table}[t]
\centering
\begin{tabular}{lcccc}
\toprule
\multicolumn{1}{c}{\multirow{2}{*}{\textbf{Model}}} & \multicolumn{2}{c}{\textbf{\taskTwo}} & \multicolumn{2}{c}{\textbf{\taskThree}} \\
\multicolumn{1}{c}{} & \textbf{MCQ} & \textbf{Y/N} & \textbf{MCQ} & \textbf{Y/N} \\
\midrule
\rowcolor{gray!20} \multicolumn{5}{c}{\textbf{Closed Source}} \\
GPT-4o & \underline{67.2} & 59.8 & \underline{80.3} & \underline{63.7} \\
Gemini 1.5 Pro & 60.4 & \underline{64.7} & 71.5 & 57.0 \\
\rowcolor{gray!20} \multicolumn{5}{c}{\textbf{Open Source}} \\
VideoChat2 & 31.3 & 63.9 & 53.8 & 51.3 \\
VideoLLaMA 2 & 53.8 & 61.3 & 53.5 & 56.1 \\
VILA-1.5 & 53.3 & 58.7 & 57.0 & 52.4 \\
LLaVA-Video & 57.4 & 61.6 & 68.7 & 58.9 \\
\bottomrule
\end{tabular}
\caption{\textbf{Validation Set results on MCQ and Y/N variants of \taskOne and \taskTwo.} The best result for each task is shown in bold, and the best model result is underlined.}
\label{tab:discval}
\end{table}

\setlength{\tabcolsep}{3pt} 
\begin{table}[t]
\centering
\begin{tabular}{lcccc}
\toprule
\multicolumn{1}{c}{\multirow{2}{*}{\textbf{Model}}} & \multicolumn{2}{c}{\textbf{\taskTwo}} & \multicolumn{2}{c}{\textbf{\taskThree}} \\
\multicolumn{1}{c}{} & \textbf{MCQ} & \textbf{Y/N} & \textbf{MCQ} & \textbf{Y/N} \\
\midrule
\rowcolor{gray!20} \multicolumn{5}{c}{\textbf{Closed Source}} \\
GPT-4o & \underline{63.2} & \underline{65.7} & \underline{78.5} & \underline{55.9} \\
Gemini 1.5 Pro & 57.1 & 62.0 & 70.6 & 49.1 \\
\rowcolor{gray!20} \multicolumn{5}{c}{\textbf{Open Source}} \\
VideoChat2 & 28.5 & 63.8 & 49.7 & 42.3 \\
VideoLLaMA 2 & 53.3 & 56.6 & 53.0 & 52.1 \\
VILA-1.5 & 49.4 & 56.5 & 52.2 & 48.7 \\
LLaVA-Video & 54.5 & 57.1 & 70.4 & 51.2 \\
\bottomrule
\end{tabular}
\caption{\textbf{Test Set results on MCQ and Y/N variants of \taskOne and \taskTwo.} The best result for each task is shown in bold, and the best model result is underlined.}
\label{tab:disctest}
\end{table}

\setlength{\tabcolsep}{2pt}
\begin{table}[ht]
  \centering
  \small 
  \begin{tabular}{lcccccc}
    \toprule
    \multicolumn{1}{c}{\textbf{Model}} & \multicolumn{2}{c}{\textbf{\taskOne}} & \multicolumn{2}{c}{\textbf{\taskTwo}} & \multicolumn{2}{c}{\textbf{\taskThree}} \\
    \cmidrule(lr){2-3} \cmidrule(lr){4-5} \cmidrule(lr){6-7}
    & \textbf{CLIP} & \textbf{LLM-M} & \textbf{CLIP} & \textbf{LLM-M} & \textbf{CLIP} & \textbf{LLM-M} \\
    \midrule
    \rowcolor{gray!20} \multicolumn{7}{c}{\textbf{Closed Source}} \\
    GPT-4o & 0.77 & 1.63 & 0.78 & 2.07 & 0.65 & 3.08 \\
    Gemini 1.5 Pro & 0.79 & 1.64 & 0.74 & 2.24 & 0.53 & 2.55 \\
    \rowcolor{gray!20} \multicolumn{7}{c}{\textbf{Open Source}} \\
    VideoChat2 & 0.71 & 1.17 & 0.68 & 1.81 & 0.62 & 2.16 \\
    VideoLLaMA 2 & 0.78 & 1.45 & 0.61 & 1.16 & 0.66 & 2.13 \\
    VILA-1.5 & 0.66 & 1.53 & 0.50 & 2.16 & 0.40 & 2.07 \\
    LLaVA-V-7B & 0.65 & 1.66 & 0.59 & 1.71 & 0.58 & 2.44 \\
    \bottomrule
  \end{tabular}
  \caption{\textbf{Validation Set Results on Gen variants.} Metrics are grouped by tasks, each with CLIP and LLM-M scores.}
  \label{tab:gen:test}
\end{table}

\setlength{\tabcolsep}{2pt}
\begin{table}[ht]
  \centering
  \small 
  \begin{tabular}{lcccccc}
    \toprule
    \multicolumn{1}{c}{\textbf{Model}} & \multicolumn{2}{c}{\textbf{\taskOne}} & \multicolumn{2}{c}{\textbf{\taskTwo}} & \multicolumn{2}{c}{\textbf{\taskThree}} \\
    \cmidrule(lr){2-3} \cmidrule(lr){4-5} \cmidrule(lr){6-7}
    & \textbf{CLIP} & \textbf{LLM-M} & \textbf{CLIP} & \textbf{LLM-M} & \textbf{CLIP} & \textbf{LLM-M} \\
    \midrule
    \rowcolor{gray!20} \multicolumn{7}{c}{\textbf{Closed Source}} \\
    GPT-4o & 0.76 & 1.66 & 0.77 & 2.09 & 0.64 & 2.92 \\
    Gemini 1.5 Pro & 0.77 & 1.57 & 0.72 & 2.05 & 0.48 & 2.45 \\
    \rowcolor{gray!20} \multicolumn{7}{c}{\textbf{Open Source}} \\
    VideoChat2 & 0.70 & 1.62 & 0.68 & 1.98 & 0.60 & 2.08 \\
    VideoLLaMA 2 & 0.77 & 1.46 & 0.62 & 1.19 & 0.65 & 1.97 \\
    VILA-1.5 & 0.66 & 1.48 & 0.49 & 2.20 & 0.39 & 2.09 \\
    LLaVA-V-7B & 0.64 & 1.50 & 0.58 & 1.70 & 0.57 & 2.29 \\
    \bottomrule
  \end{tabular}
  \caption{\textbf{Test Set Results on Gen variants.} Metrics are grouped by tasks, each with CLIP and LLM-M scores.}
  \label{tab:gen:test}
\end{table}

We release our data with two splits: a validation split, where we make ground truth labels accessible, and a test split, where we hide ground truth labels. This is to prevent misuse of our data. The validation set contains 827 videos (50\% of data), of which 95 are from the hard subset (Sec. \ref{sec:analysis:hard_set}). The test set contains 828 videos, of which 224 are from the hard subset. This makes the test set slightly more challenging than the validation set.

Results on MCQ and Y/N variants for the validation and test sets are in Tables \ref{tab:discval} and \ref{tab:disctest} respectively. 

\section{Analysis}
\label{app:Analysis}

\subsection{Perception-Comprehension}
\label{app:Analysis:PC}

We collect human perception and comprehension descriptions using the templates in Figure \ref{fig:pc:p} and Figure \ref{fig:pc:c} respectively. Once again, we use the same location, age and education criterion to select annotators as before. We pay \$0.26 per annotation.

For perception input to LLaVA-Video, we modify the prompt for \taskTwo--MCQ to:
\begin{prompt}
    The beginning of the video shows <p\_preevent>. The end of the video shows <p\_postevent>. \\
    
    Which of the following descriptions indicate what happened in the hidden (black) frames of the video? A. <Option A> B. <Option B> C. <Option C>
\end{prompt}

For perception and comprehension, we further modify the prompt to:
\begin{prompt}
    The beginning of the video shows <p\_preevent>. The end of the video shows <p\_postevent>. The two parts differ in the following way: <pc\_comp>. \\
    
    Which of the following descriptions indicate what happened in the hidden (black) frames of the video? A. <Option A> B. <Option B> C. <Option C>
\end{prompt}

For examples with full video clips, please watch the video \texttt{examples.mp4} in the supplementary zip file. In Figure \ref{fig:pcegs}, we show an example how perception and comprehension have helped the baseline model arrive at correct answers.


\subsection{CoT examples}
\label{app:Analysis:CoTquali}

\begin{figure*} 
    \centering

    \begin{subfigure}[b]{1\textwidth} 
        \centering
        \includegraphics[width=\textwidth, page=1]{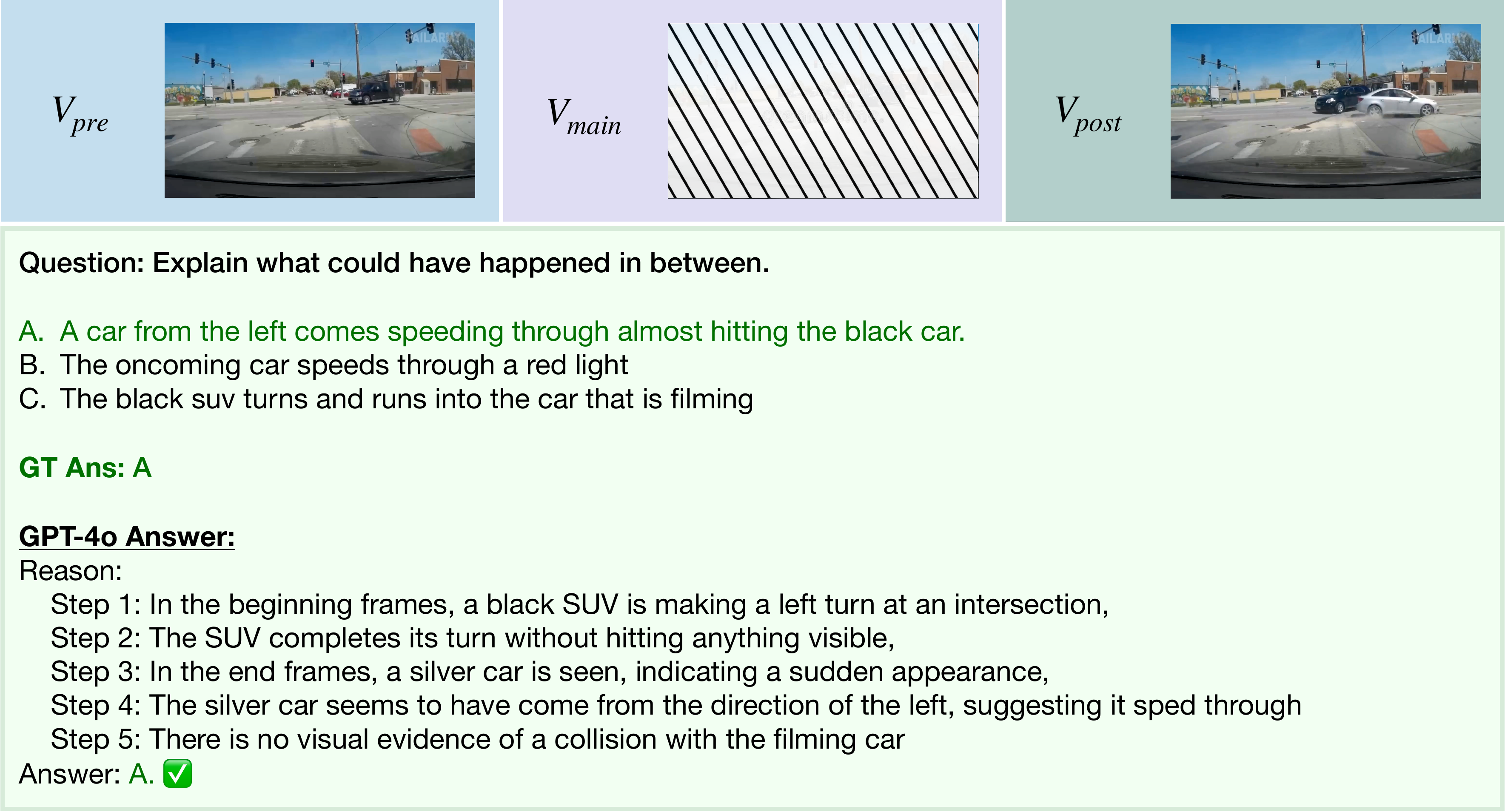}
        \caption{Example where GPT-4o gets it right}
    \end{subfigure}
    \vskip\baselineskip

    \begin{subfigure}[b]{1\textwidth}
        \centering
        \includegraphics[width=\textwidth, page=2]{supp-fig/VARFigures-egs-crop.pdf}
        \caption{Example where GPT-4o gets it wrong}
    \end{subfigure}
    \vskip\baselineskip


    \caption{\textbf{CoT Examples with GPT-4o.} (a) has a case where GPT-4o shows good step-by-step reasoning to arrive at the correct answer. (b) shows that GPT-4o sometimes relies of preconceived notions in its reasoning steps, and therefore makes a mistake.}
    \label{fig:cotegs}
\end{figure*}

Because of the low-quality of LLaVA-Video on this task, we focus on GPT-4o. In Figure \ref{fig:cotegs}, we show examples of how step-by-step reasoning affects answer choice selection for GPT-4o. In many cases, these reasoning steps are correct (\eg in part (a)). On the other hand, preconceptions about the scene can influence the reasoning process, leading to mistakes (\eg in part (b)). Again, for examples with full video clips, please watch the video \texttt{examples.mp4} in the supplementary zip file.

\subsection{Challenging subset examples}
\label{app:Analysis:hardsubsetegs}

Hard-subset examples with video clips are shown in \texttt{examples.mp4} in the supplementary zip file. These are marked with an orange asterisk on the top right corner.

\section{Limitations and Ethical Considerations}
\label{app:Limitations}
\noindent\textbf{Data Source}: As we use videos from the test set of Oops! \cite{epstein2020oops}, it is possible that some VLMs trained on extensive datasets may already be familiar with similar content. However, our benchmark can be extended to include newer videos in future.

\noindent\textbf{Evaluation Metrics}: Although we defined our MCQ and Y/N tasks to challenge models to perform more complex reasoning, quantitative metrics like accuracy might not reflect the depth of reasoning or the logical processes involved. For generative tasks, current metrics struggle to capture the nuances involved. We address this by asking humans to evaluate the thoughtfulness and visual contradictions in reasoning, but further research is needed to automatically evaluate explanations generated by models. Furthermore, given the open-ended nature of our generative tasks, our human evaluation may be subject to bias. Our MCQ and Y/N variants, on the other hand, do not suffer from this limitation. Finally, we conduct all our human annotations according to ethics policies.

\noindent\textbf{Pre-training strategies}: The models evaluated are primarily trained on language modeling, and may not have been explicitly trained for abductive and defeasible reasoning, potentially limiting their performance. Further research is required to study how different pretraining or finetuning approaches may perform on this task.

\noindent\textbf{Explanation complexity}: Our current annotation process focuses on free-form explanations for the unexpected scenarios, however, it may be interesting to study how more scientific reasoning (e.g., using intuitive physics to explain a fall) may influence the performance of models.

\noindent\textbf{Size and diversity}: Although \bench{} includes 15,469 questions, the size and diversity might still be insufficient to generalize findings across all types of reasoning challenges and rare and highly uncommon events might still be underrepresented.





\begin{figure*}[p] 
    \centering

    \begin{subfigure}[b]{1\textwidth} 
        \centering
        \includegraphics[width=\textwidth, page=4]{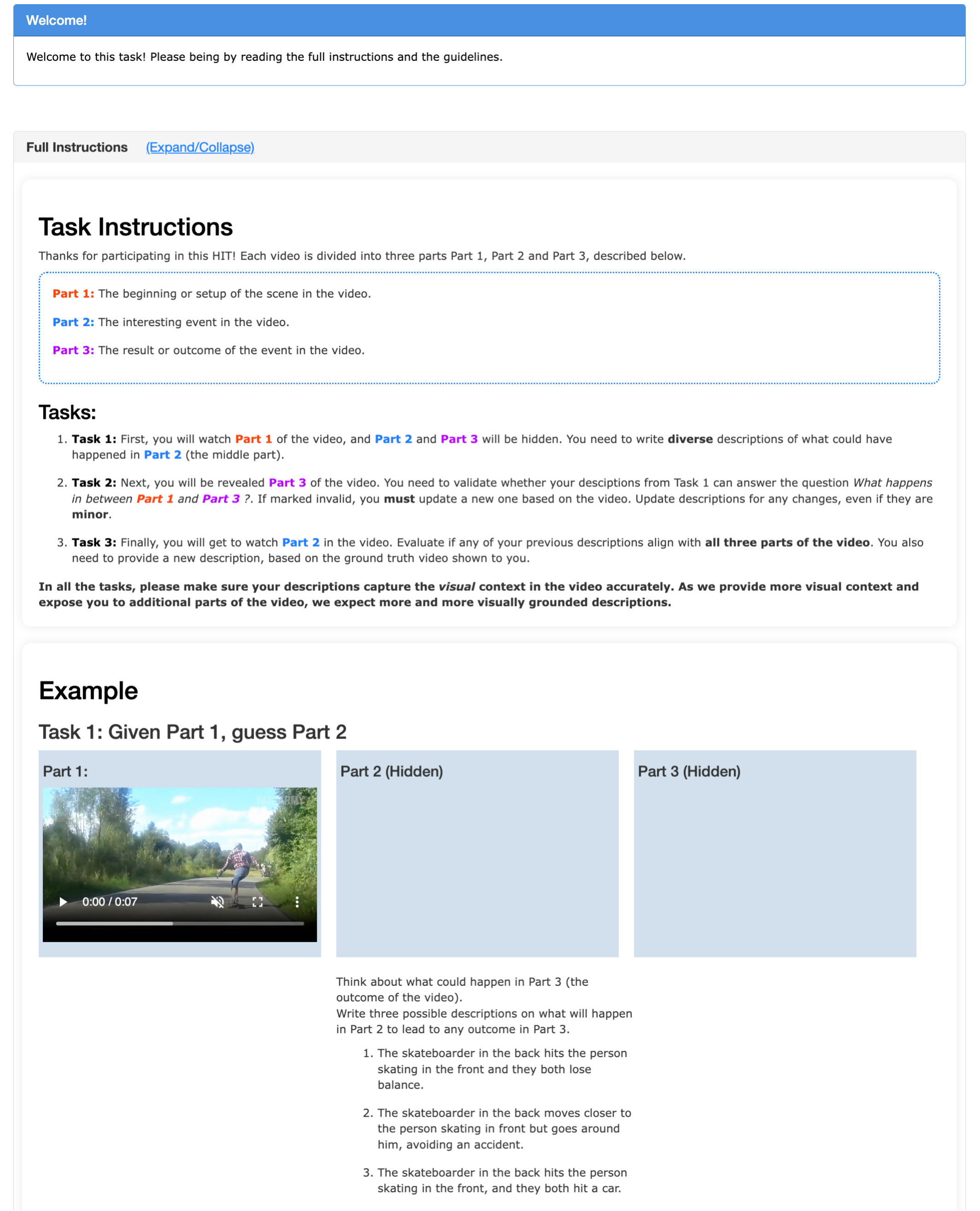}
        \caption{1 of 6}
    \end{subfigure}
    \caption{Black Swan Annotation Template}
    \label{fig:mainannottemplate}
\end{figure*}

\begin{figure*}[p]
    \ContinuedFloat
    \centering

    \begin{subfigure}[b]{1\textwidth}
        \centering
        \includegraphics[width=0.9\textwidth, page=5]{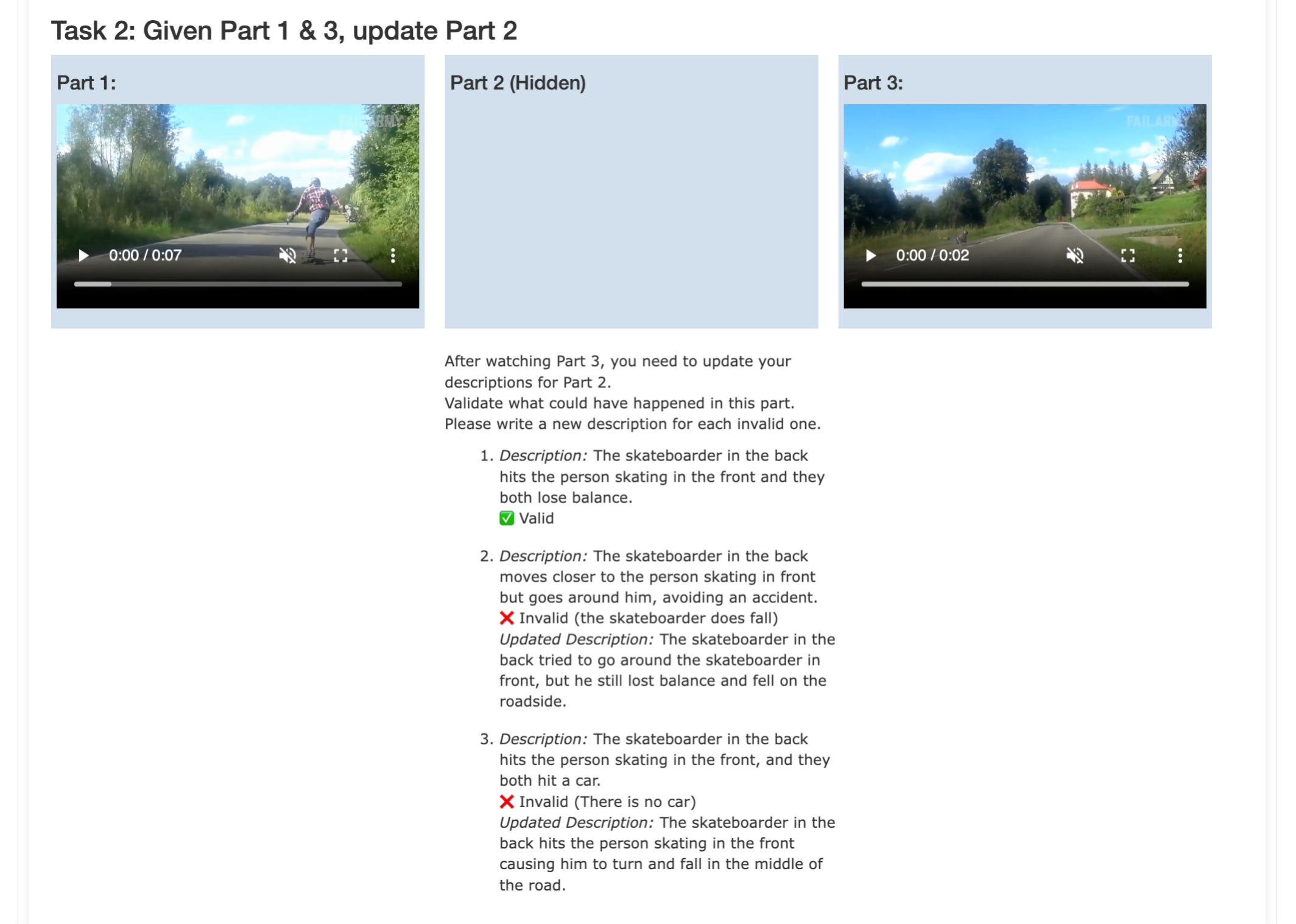}
        \caption{2 of 6}
    \end{subfigure}
    \caption{(Continued) Black Swan Annotation Template}
\end{figure*}

\begin{figure*}[p]
    \ContinuedFloat
    \centering

    \begin{subfigure}[b]{1\textwidth}
        \centering
        \includegraphics[width=0.9\textwidth, page=6]{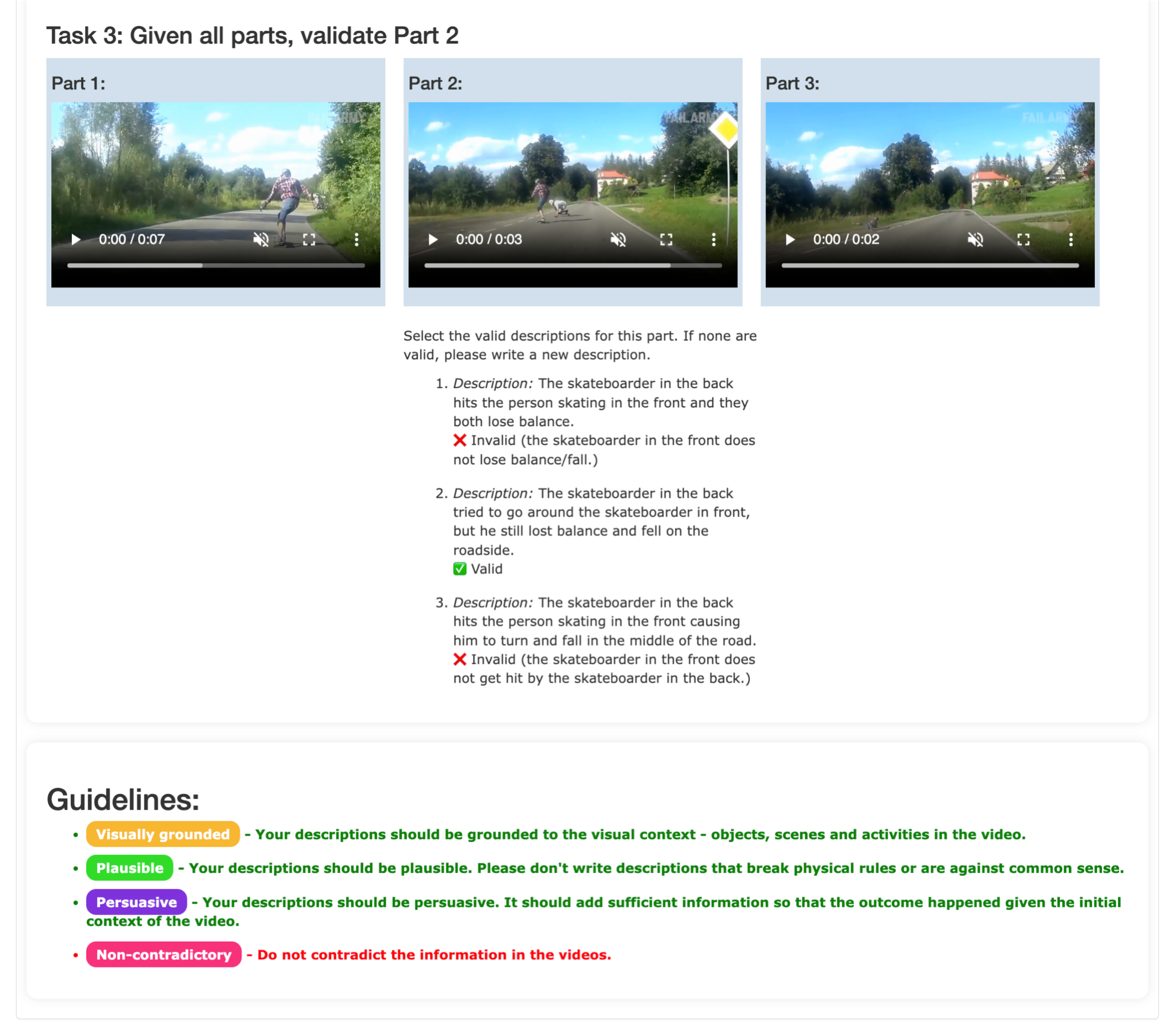}
        \caption{3 of 6}
    \end{subfigure}
    \caption{(Continued) Black Swan Annotation Template}
\end{figure*}

\begin{figure*}[p]
    \ContinuedFloat
    \centering

    \begin{subfigure}[b]{1\textwidth}
        \centering
        \includegraphics[width=0.9\textwidth, page=7]{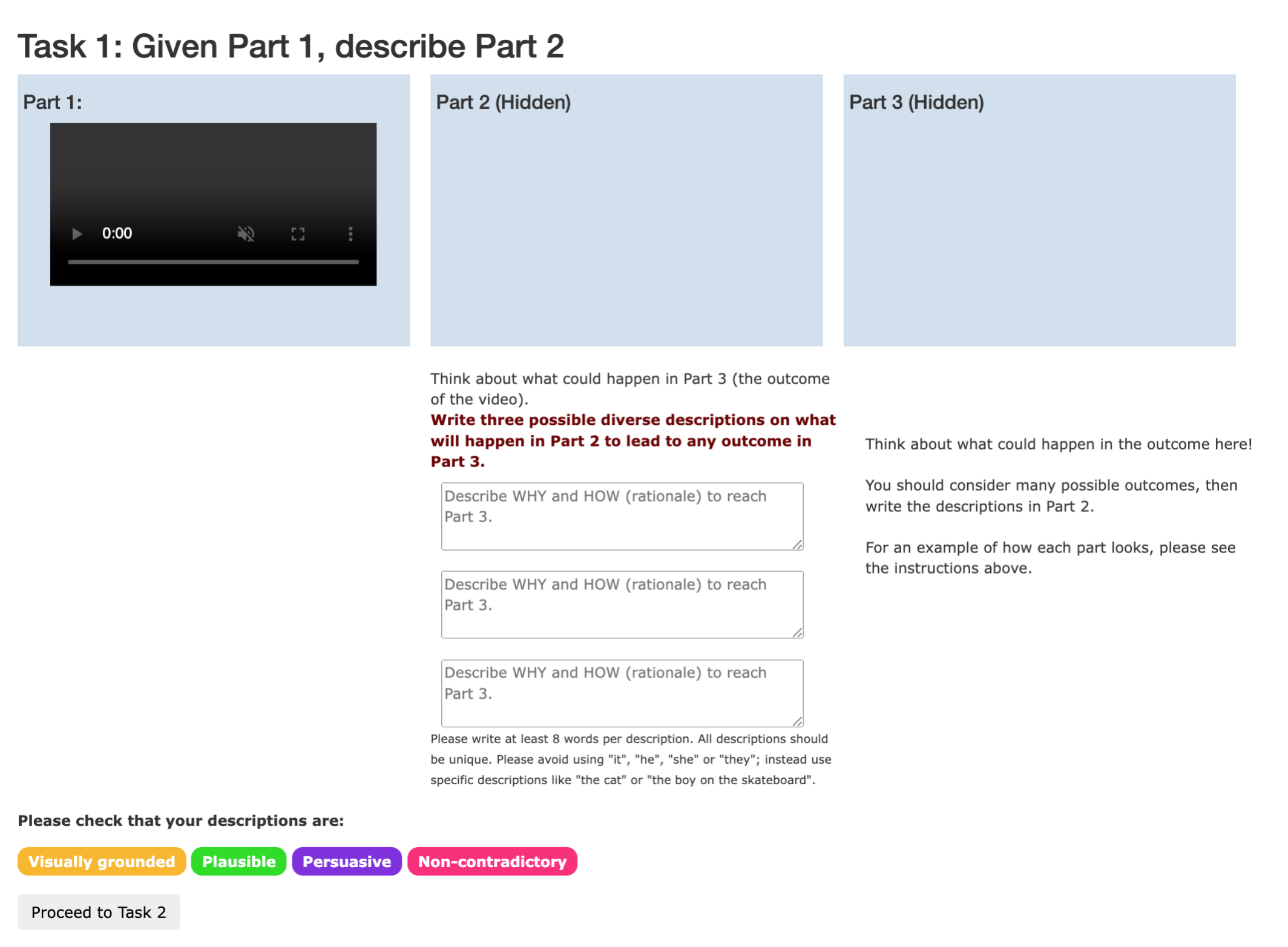}
        \caption{4 of 6}
    \end{subfigure}
    \caption{(Continued) Black Swan Annotation Template}
\end{figure*}

\begin{figure*}[p]
    \ContinuedFloat
    \centering

    \begin{subfigure}[b]{1\textwidth}
        \centering
        \includegraphics[width=0.9\textwidth, page=8]{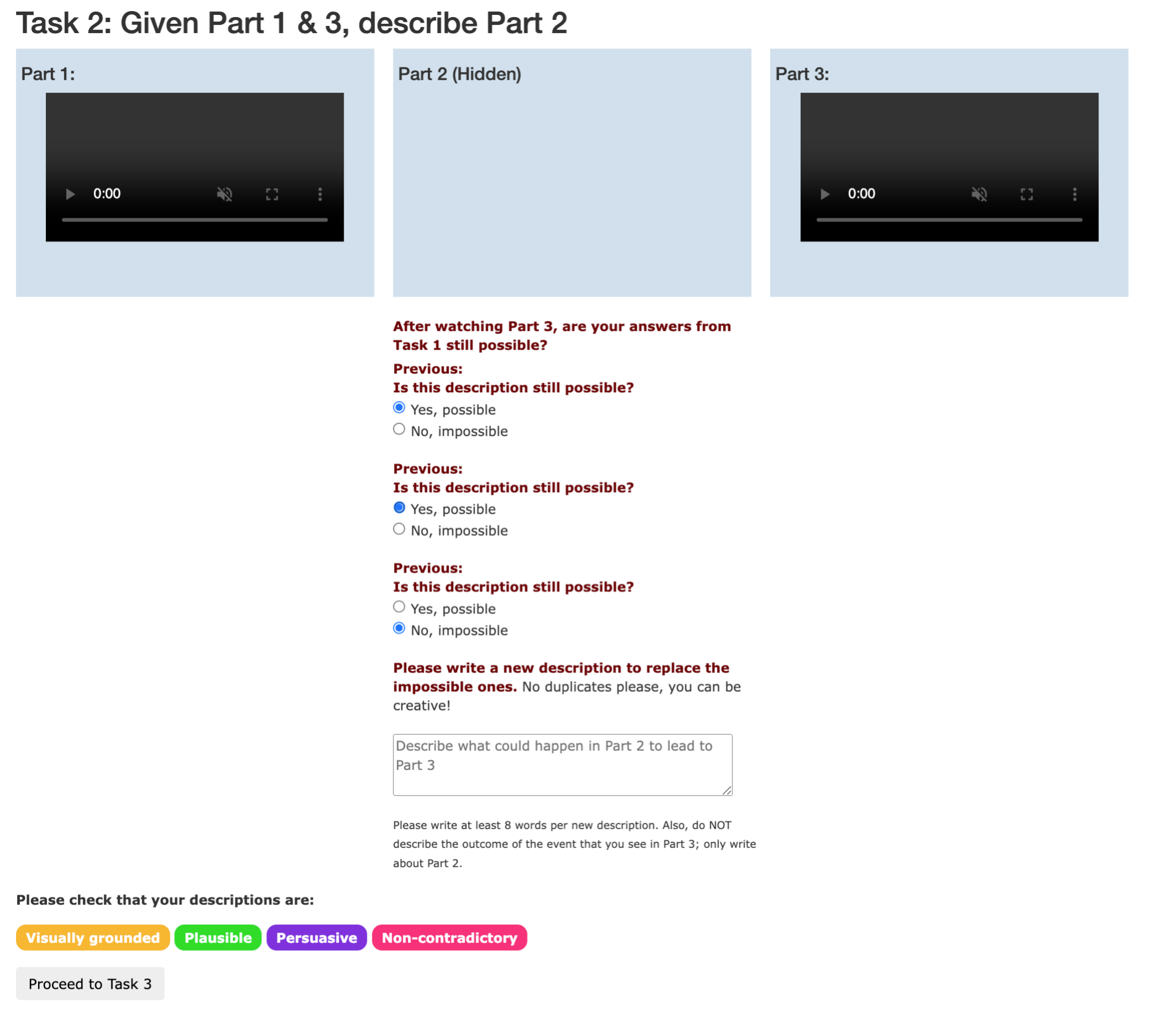}
        \caption{5 of 6}
    \end{subfigure}
    \caption{(Continued) Black Swan Annotation Template}
\end{figure*}

\begin{figure*}[p]
    \ContinuedFloat
    \centering

    \begin{subfigure}[b]{1\textwidth}
        \centering
        \includegraphics[width=0.9\textwidth, page=9]{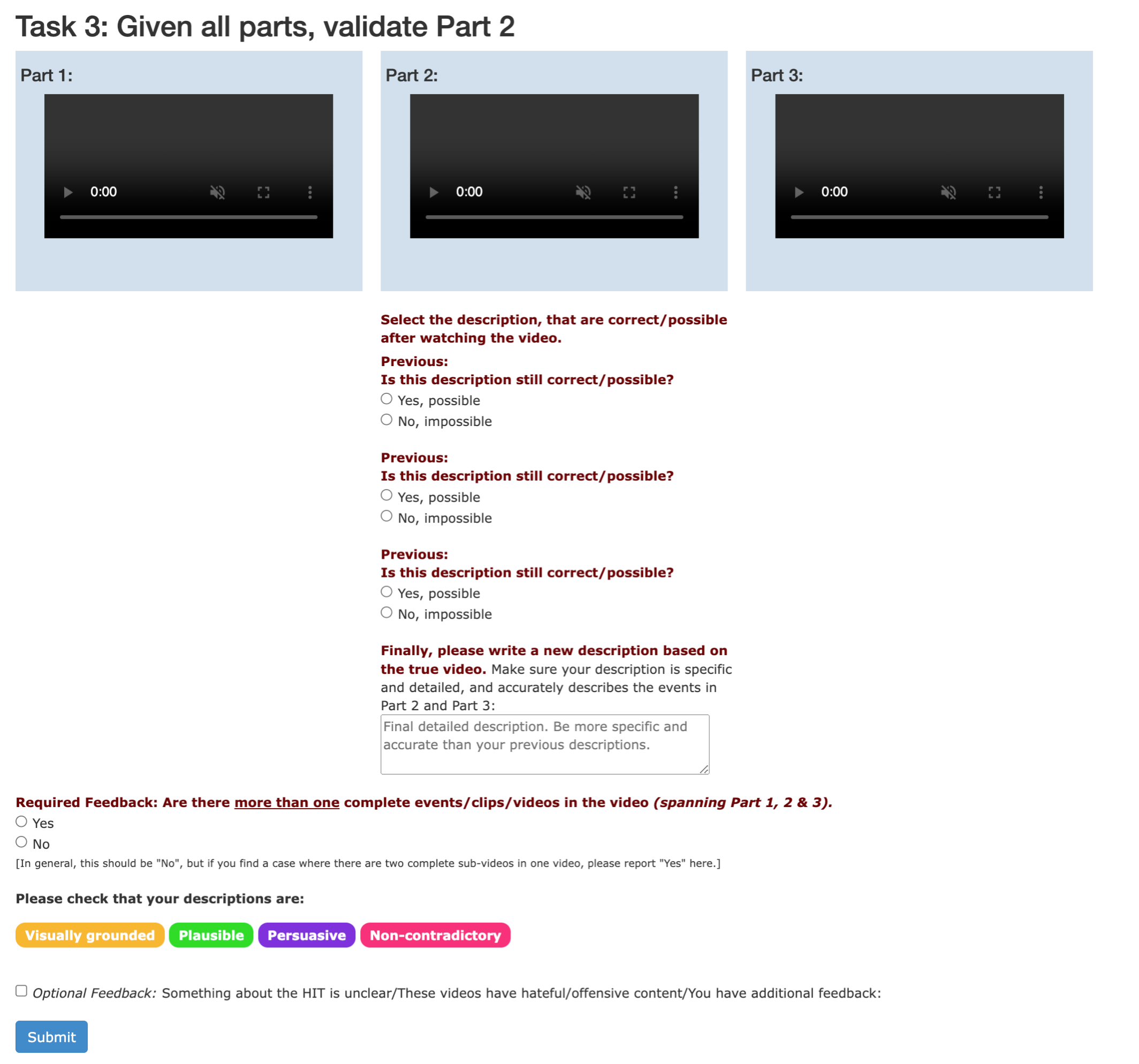}
        \caption{6 of 6}
    \end{subfigure}
    \caption{(Continued) Black Swan Annotation Template}
\end{figure*}
\begin{figure*}[p] 
    \centering

    \begin{subfigure}[b]{1\textwidth} 
        \centering
        \includegraphics[width=\textwidth, page=1]{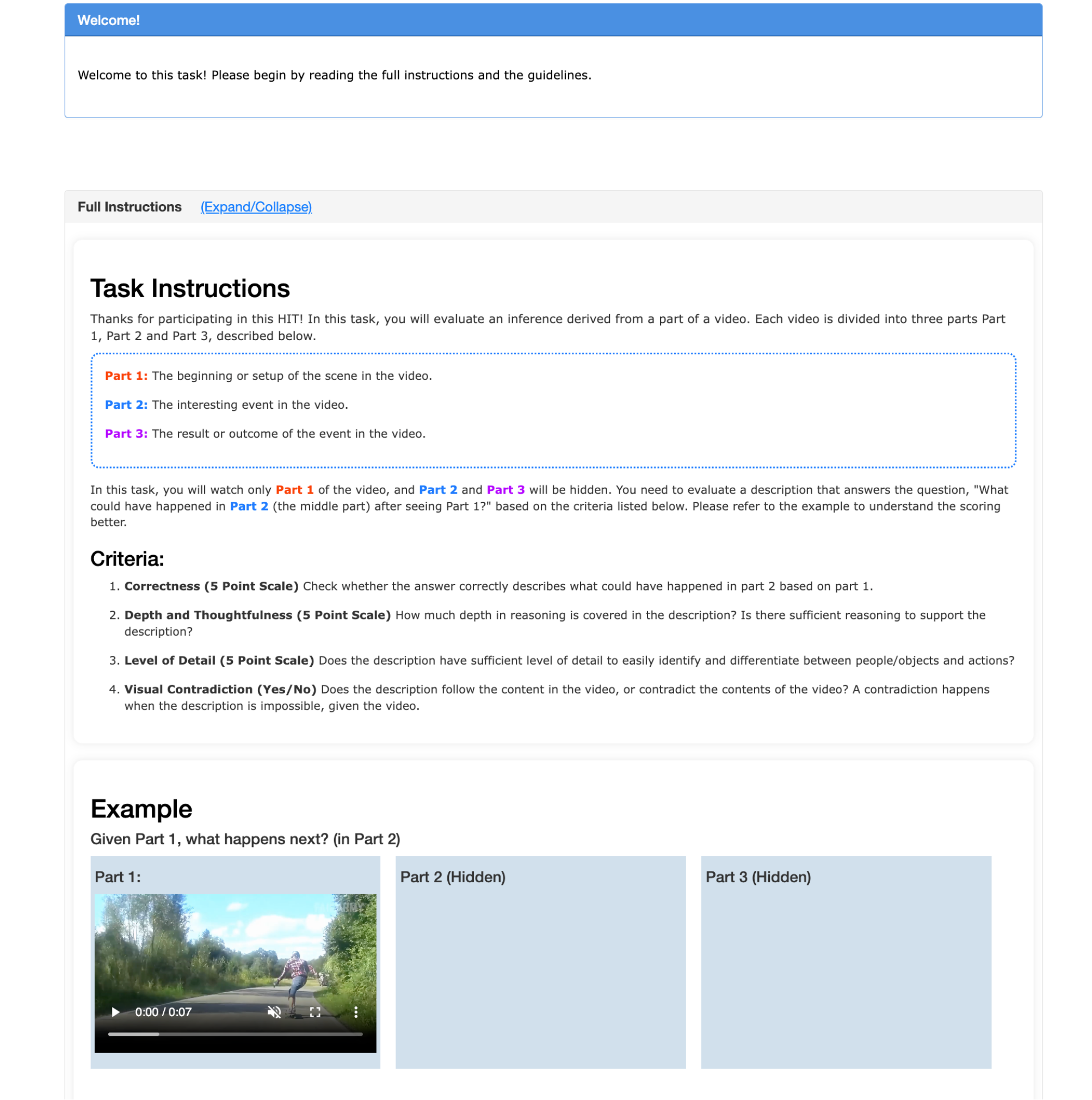}
        \caption{1 of 3}
    \end{subfigure}
    \caption{Human Evaluation for Generative Tasks}
    \label{fig:humanevalgen}
\end{figure*}

\begin{figure*}[p]
    \ContinuedFloat
    \centering

    \begin{subfigure}[b]{1\textwidth}
        \centering
        \includegraphics[width=0.9\textwidth, page=2]{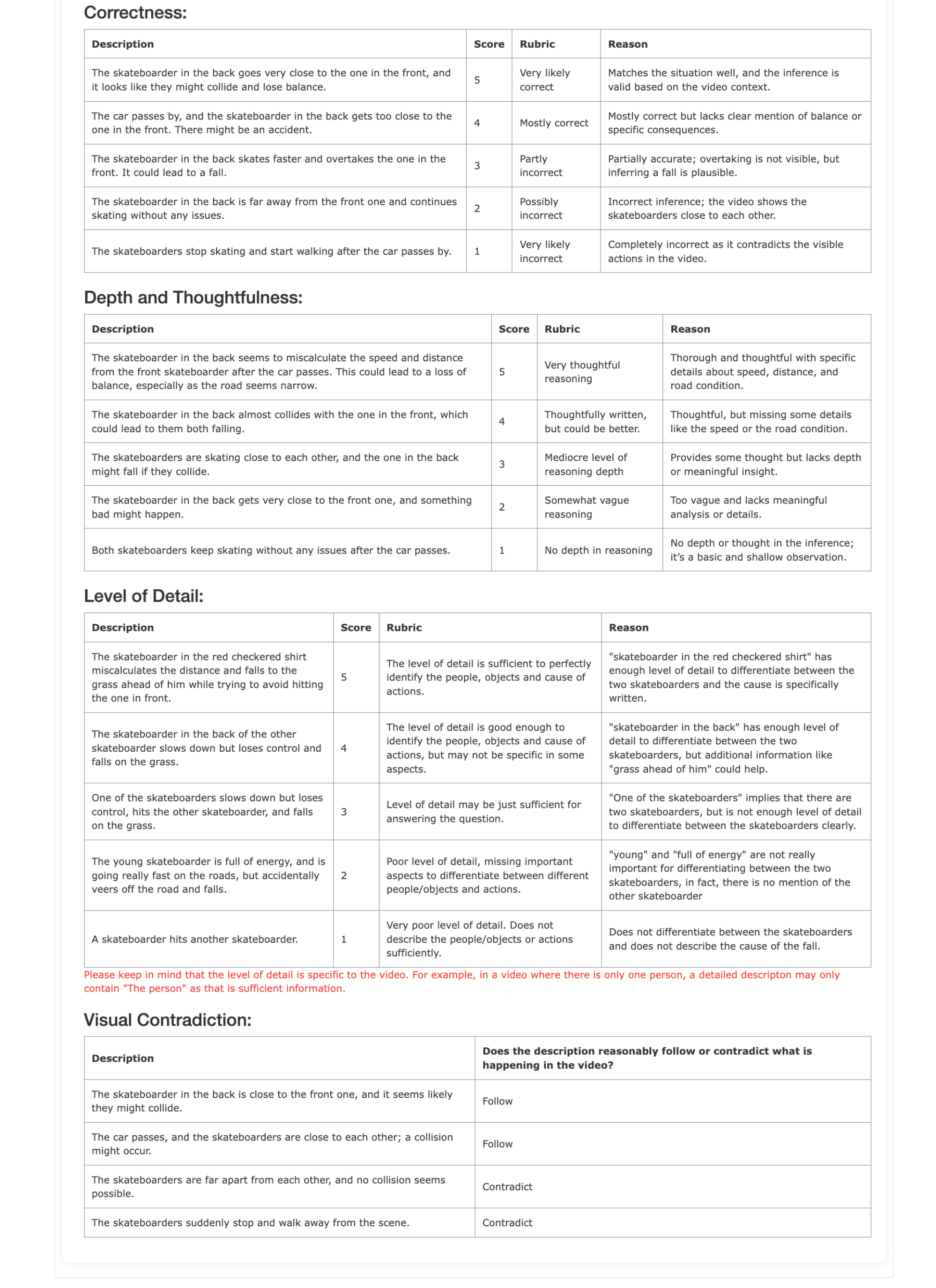}
        \caption{2 of 3}
    \end{subfigure}
    \caption{(Continued) Human Evaluation for Generative Tasks}
\end{figure*}

\begin{figure*}[p]
    \ContinuedFloat
    \centering

    \begin{subfigure}[b]{1\textwidth}
        \centering
        \includegraphics[width=0.9\textwidth, page=3]{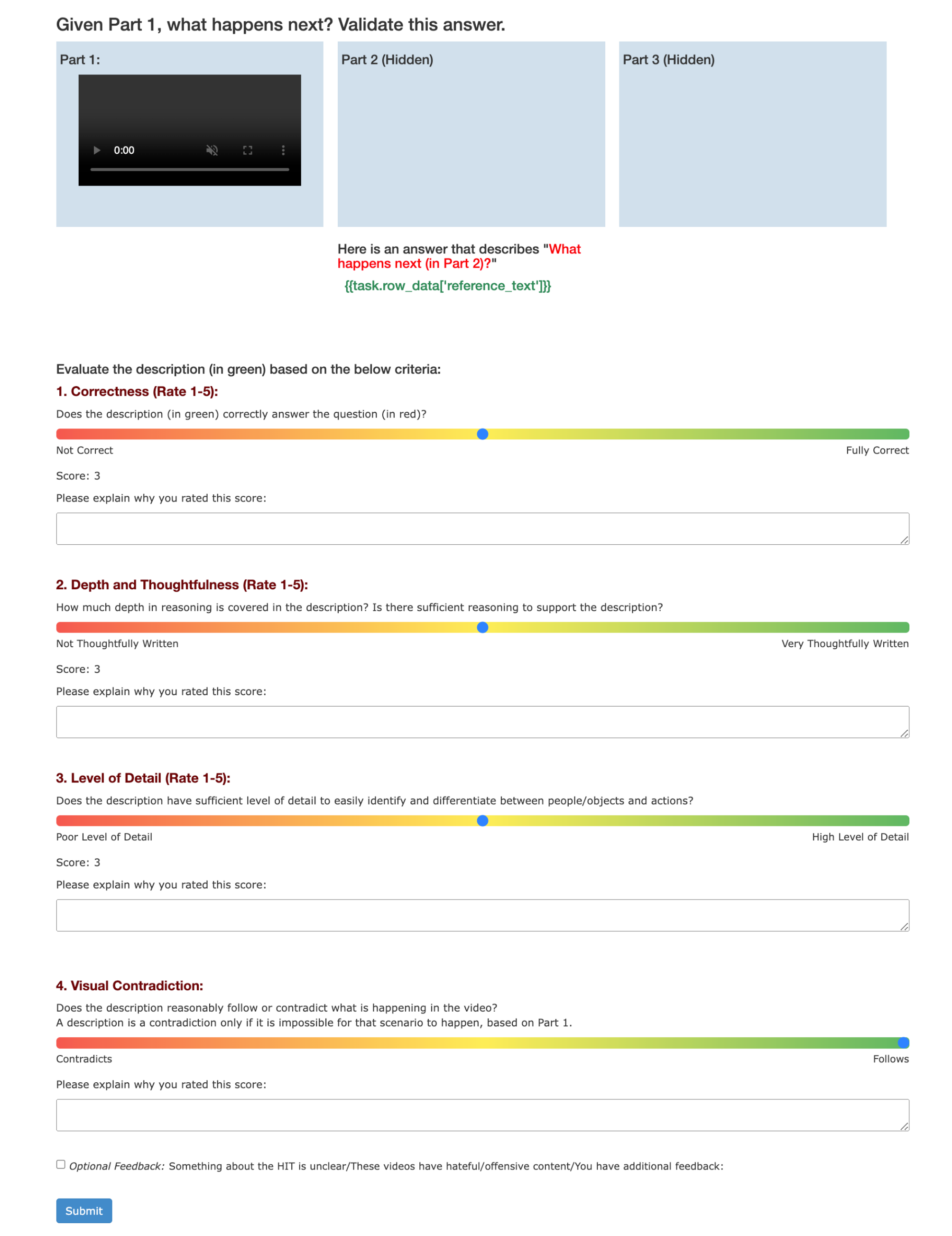}
        \caption{3 of 3}
    \end{subfigure}
    \caption{(Continued) Human Evaluation for Generative Tasks}
\end{figure*}
\begin{figure*}[p] 
    \centering

    \begin{subfigure}[b]{1\textwidth} 
        \centering
        \includegraphics[width=\textwidth, page=1]{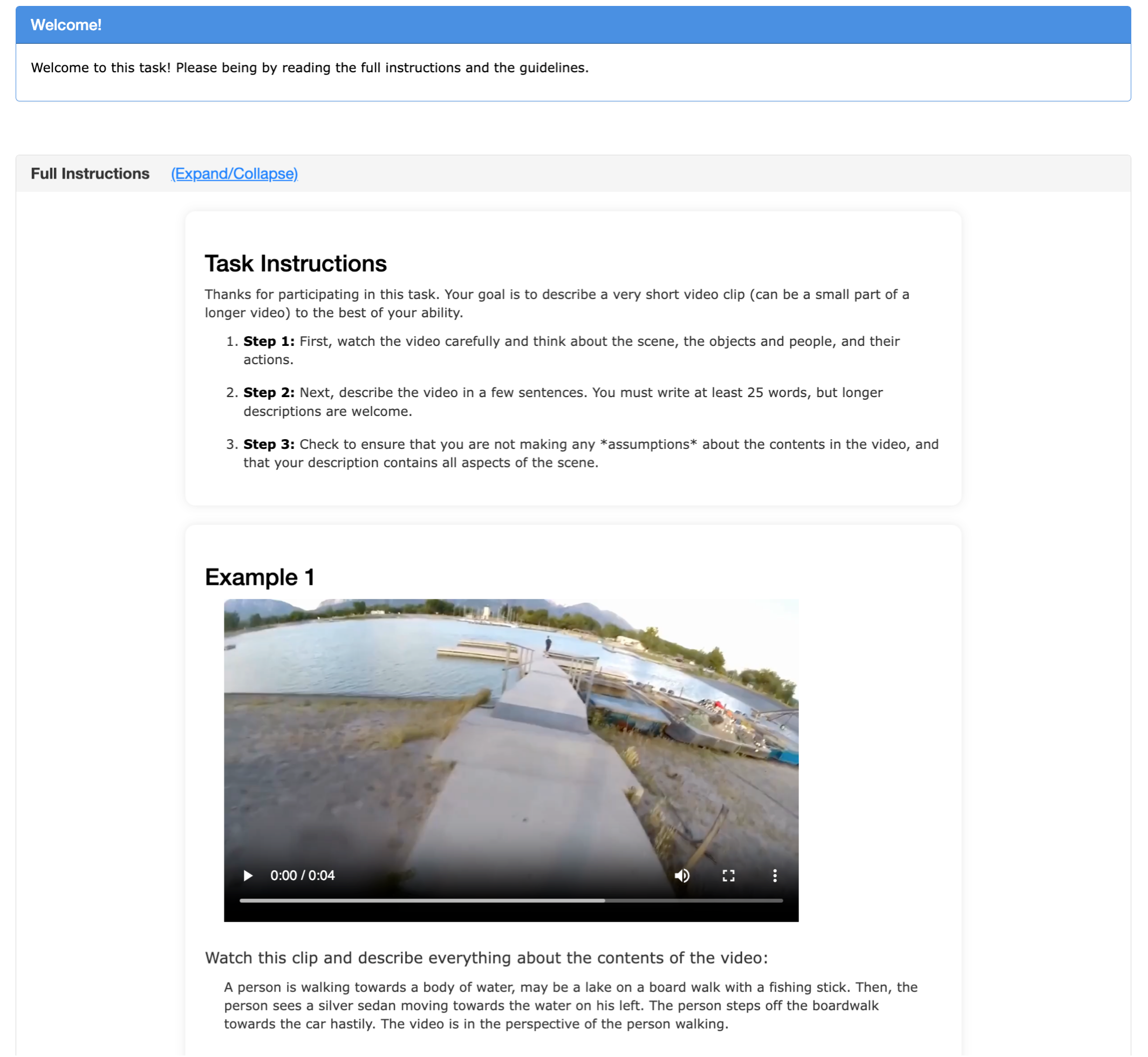}
        \caption{1 of 3}
    \end{subfigure}
    \caption{Analysis: Collection of Human Perception}
    \label{fig:pc:p}
\end{figure*}

\begin{figure*}[p]
    \ContinuedFloat
    \centering

    \begin{subfigure}[b]{1\textwidth}
        \centering
        \includegraphics[width=0.9\textwidth, page=2]{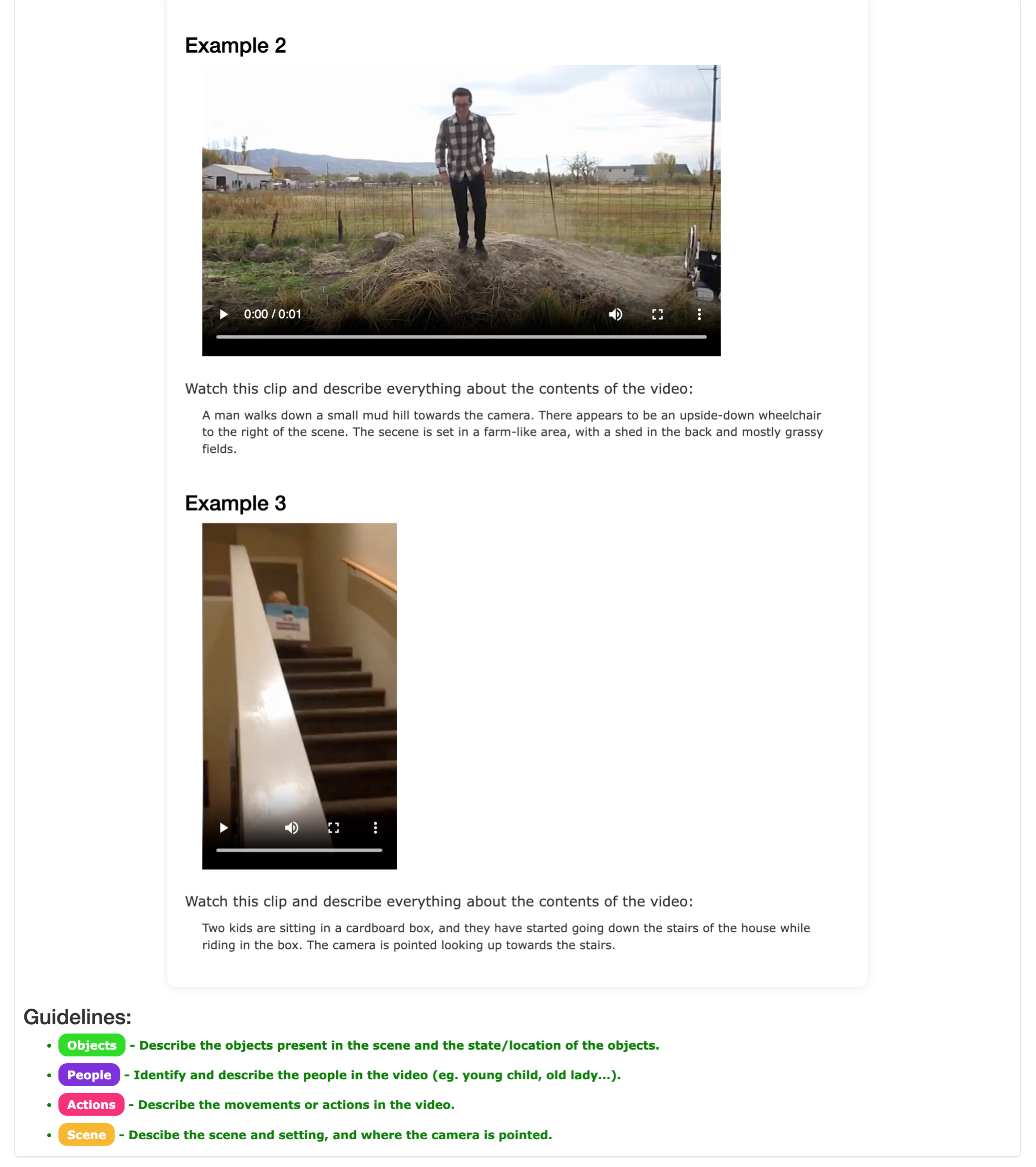}
        \caption{2 of 3}
    \end{subfigure}
    \caption{(Continued) Analysis: Collection of Human Perception}
\end{figure*}

\begin{figure*}[p]
    \ContinuedFloat
    \centering

    \begin{subfigure}[b]{1\textwidth}
        \centering
        \includegraphics[width=0.9\textwidth, page=3]{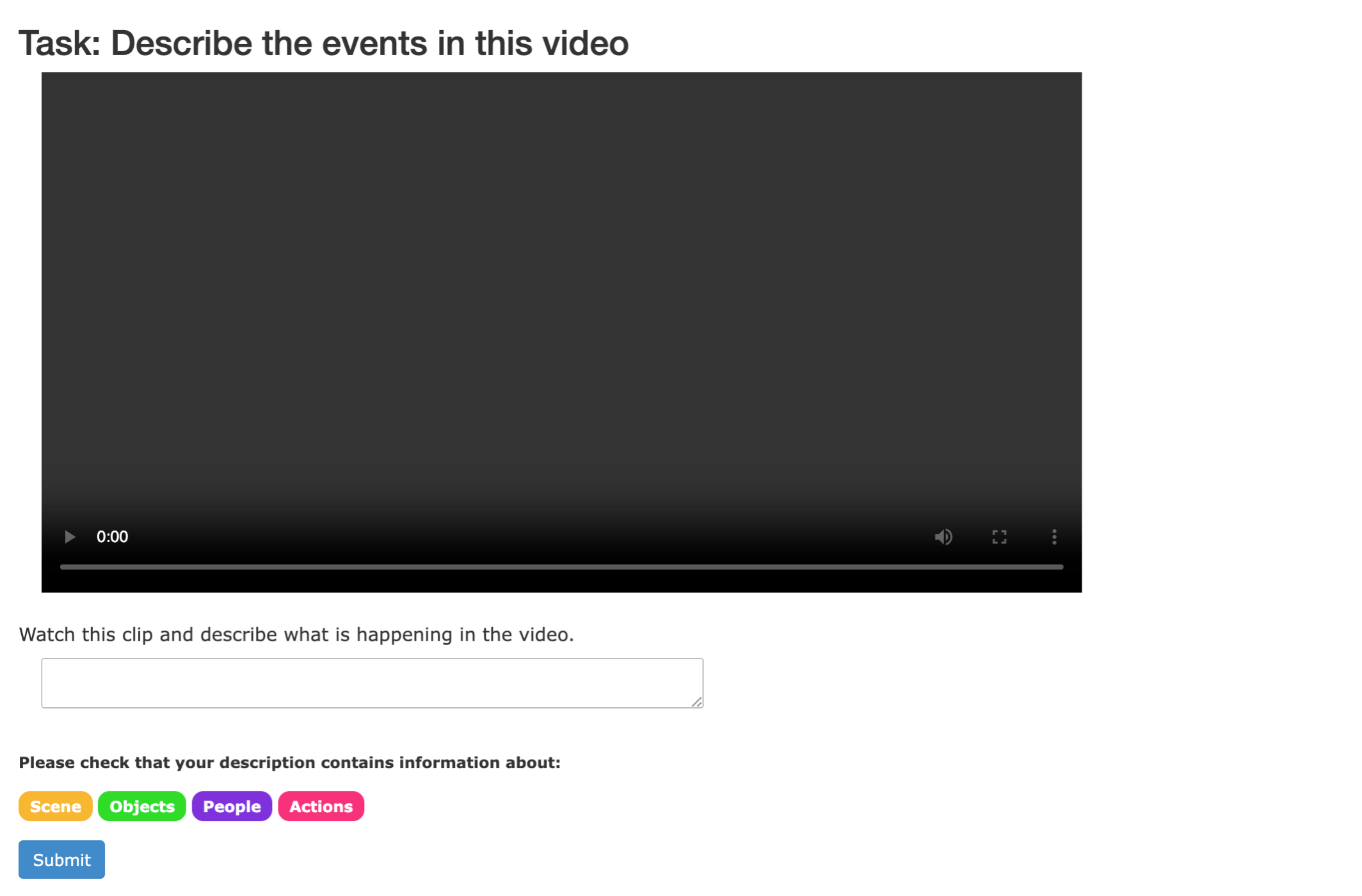}
        \caption{3 of 3}
    \end{subfigure}
    \caption{(Continued) Analysis: Collection of Human Perception}
\end{figure*}
\begin{figure*}[p] 
    \centering

    \begin{subfigure}[b]{1\textwidth} 
        \centering
        \includegraphics[width=\textwidth, page=4]{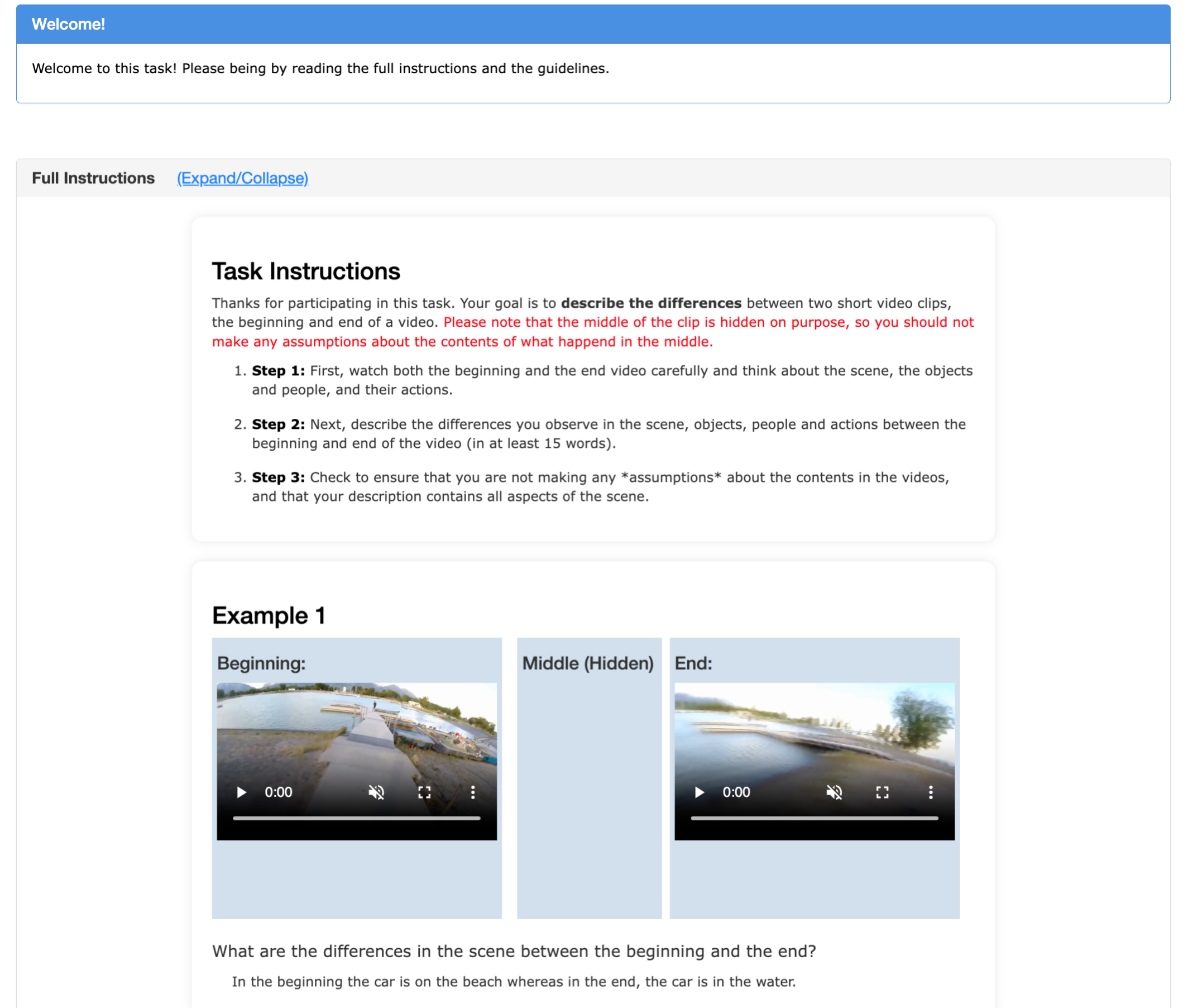}
        \caption{1 of 3}
    \end{subfigure}
    \caption{Analysis: Collection of Human Comprehension}
    \label{fig:pc:c}
\end{figure*}

\begin{figure*}[p]
    \ContinuedFloat
    \centering

    \begin{subfigure}[b]{1\textwidth}
        \centering
        \includegraphics[width=0.9\textwidth, page=5]{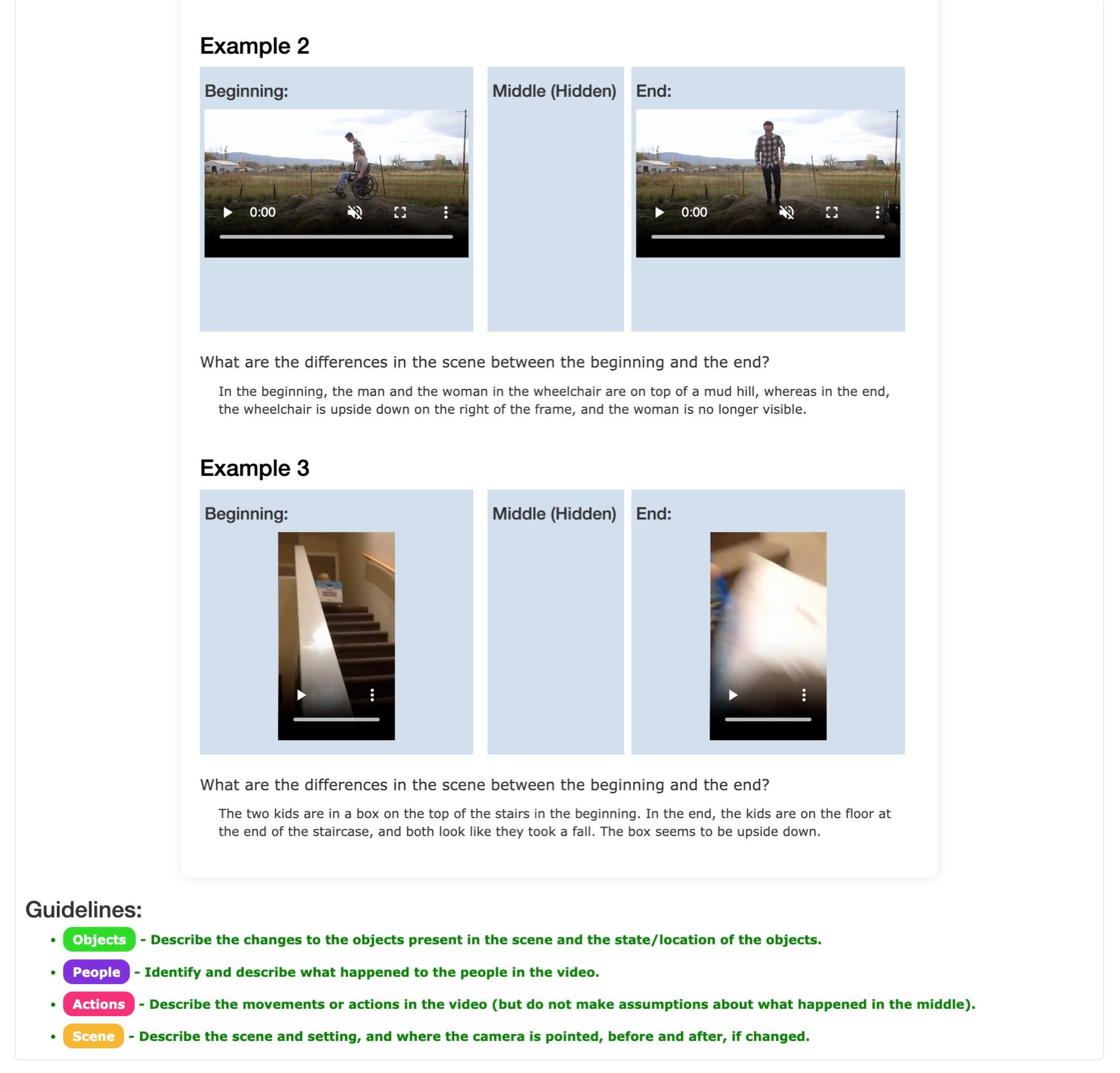}
        \caption{2 of 3}
    \end{subfigure}
    \caption{(Continued) Analysis: Collection of Human Comprehension}
\end{figure*}

\begin{figure*}[p]
    \ContinuedFloat
    \centering

    \begin{subfigure}[b]{1\textwidth}
        \centering
        \includegraphics[width=0.9\textwidth, page=6]{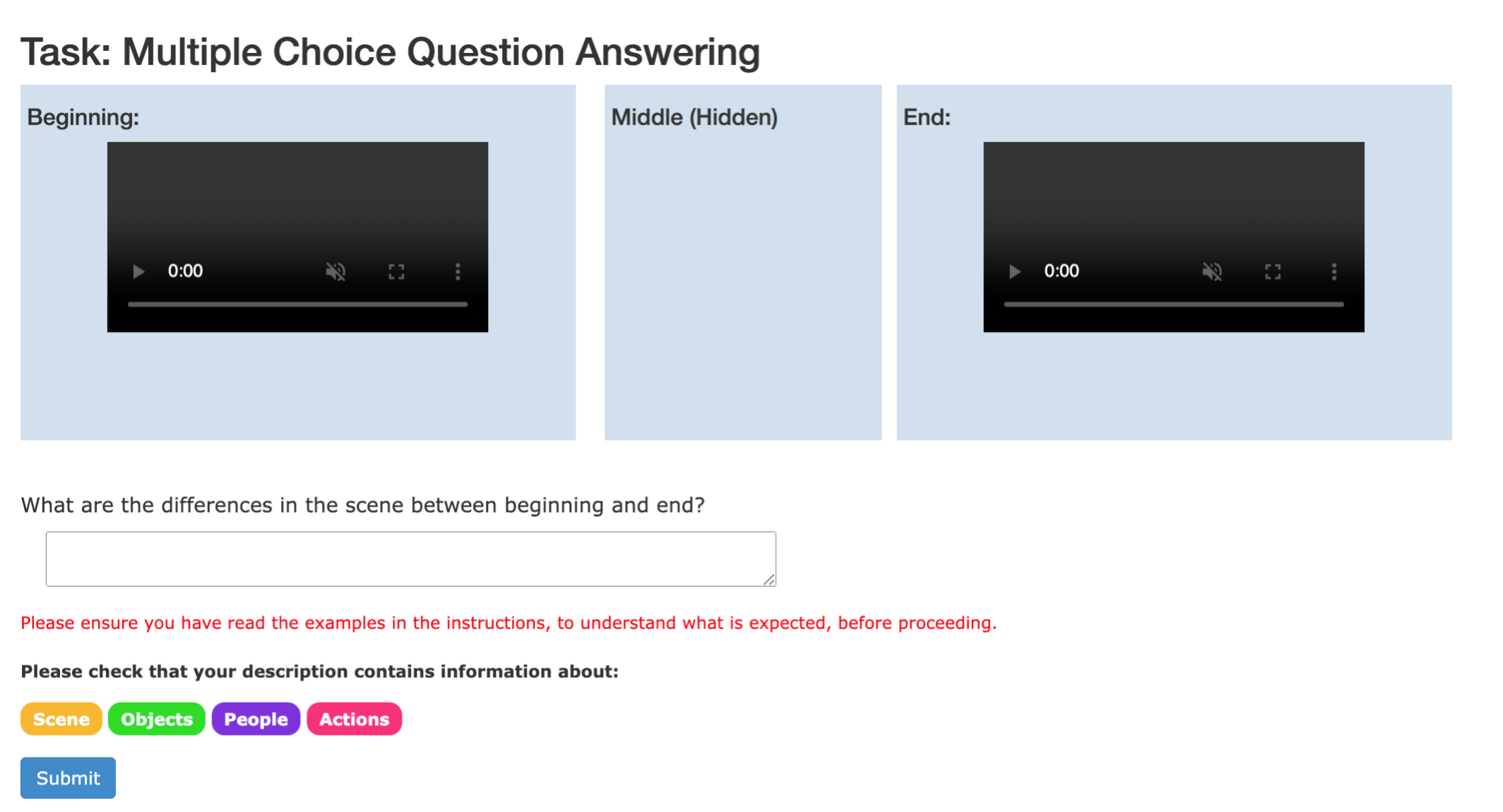}
        \caption{3 of 3}
    \end{subfigure}
    \caption{(Continued) Analysis: Collection of Human Comprehension}
\end{figure*}

\end{document}